\documentclass[sn-apa,iicol]{sn-jnl}

\usepackage{graphicx}
\usepackage{multirow}
\usepackage{amsmath,amsthm,amssymb,amsfonts}
\usepackage{mathrsfs}
\usepackage[title]{appendix}%
\usepackage{textcomp}%
\usepackage{manyfoot}%
\usepackage{algorithm}%
\usepackage{algorithmicx}%
\usepackage{algpseudocode}%
\usepackage{listings}%
\usepackage{latexsym}
\usepackage{supertabular}
\usepackage{multirow, url, booktabs,color, bm}
\usepackage{bbding} 
\usepackage[caption=false,font=footnotesize,labelfont=rm,textfont=rm]{subfig}
\usepackage{caption}
\usepackage{pifont}
\usepackage{xcolor}
\usepackage{verbatim}
\usepackage[percent]{overpic}
\usepackage{bbding}
\newcommand{\tabincell}[2]{\begin{tabular}{@{}#1@{}}#2\end{tabular}}
\usepackage{stfloats}
\usepackage{cases}

\newcommand{\revise}[1]{\color{black} #1 \color{black}}

\theoremstyle{thmstyleone}%
\theoremstyle{thmstyletwo}%

\theoremstyle{thmstylethree}%

\definecolor{somegray}{rgb}{0.5, 0.5, 0.5}
\newcommand{\darkgrayed}[1]{\textcolor{somegray}{#1}}
\makeatletter
\newcommand*\titleheader[1]{\gdef\@titleheader{#1}}
\AtBeginDocument{%
  \let\st@red@title\@title
  \def\@title{%
    \vskip-3em
    \bgroup\normalfont\large\centering\@titleheader\par\egroup
    \vskip1.5em\st@red@title}
}
\makeatother

\titleheader{\darkgrayed{This paper has been accepted for publication at the \\
International Journal of Computer Vision (IJCV), 2024.
\copyright Springer}}

\title{RGB Guided ToF Imaging System: A Survey of Deep Learning-based Methods}

\begin{document}

\author[1]{\fnm{Xin} \sur{Qiao}}

\author[2,3]{\fnm{Matteo} \sur{Poggi}}

\author[1]{\fnm{Pengchao} \sur{Deng}}

\author[1]{\fnm{Hao} \sur{Wei}}

\author*[1]{\fnm{Chenyang} \sur{Ge}}\email{cyge@mail.xjtu.edu.cn}

\author[2,3]{\fnm{Stefano} \sur{Mattoccia}}

\affil*[1]{\orgdiv{Institute of Artificial Intelligence and Robotics}, \orgname{Xi'an Jiaotong University}, \orgaddress{\street{No.28, West Xianning Road}, \city{Xi'an}, \postcode{710049}, \state{Shaanxi}, \country{China}}}

\affil[2]{\orgdiv{Department of Computer Science and Engineering (DISI)}, \orgname{University of Bologna}, \orgaddress{\street{Viale Risorgimento, 2}, \city{Bologna}, \postcode{40136}, \state{Emilia-Romagna}, \country{Italy}} \\
$^3$\orgdiv{Advanced Research Center on Electronic System (ARCES)}, \orgname{University of Bologna}, \orgaddress{\street{Via Vincenzo Toffano, 2/2}, \city{Bologna}, \postcode{40125}, \state{Emilia-Romagna}, \country{Italy}}}


\abstract{
Integrating an RGB camera into a ToF imaging system has become a significant technique for perceiving the real world. The RGB guided ToF imaging system is crucial to several applications, including face anti-spoofing, saliency detection, and trajectory prediction. Depending on the distance of the working range, the implementation schemes of the RGB guided ToF imaging systems are different. Specifically, ToF sensors with a uniform field of illumination, which can output dense depth but have low resolution, are typically used for close-range measurements. In contrast, LiDARs, which emit laser pulses and can only capture sparse depth, are usually employed for long-range detection. In the two cases, depth quality improvement for RGB guided ToF imaging corresponds to two sub-tasks: guided depth super-resolution and guided depth completion. In light of the recent significant boost to the field provided by deep learning, this paper comprehensively reviews the works related to RGB guided ToF imaging, including network structures, learning strategies, evaluation metrics, benchmark datasets, and objective functions. Besides, we present quantitative comparisons of state-of-the-art methods on widely used benchmark datasets. Finally, we discuss future trends and the challenges in real applications for further research.}


\maketitle

\section{Introduction} 
The real world is three-dimensional, yet usually recorded with 2D vision sensors such as color cameras. 
To faithfully portray it, a diffuse trend is to employ a depth camera during collection, widely used nowadays in downstream tasks related to computer vision and pattern recognition, such as face anti-spoofing~\citep{liu2021casia,deng2022multi}, saliency detection~\citep{cong2018hscs}, autonomous driving~\citep{hane20173d,wang2019pseudo} and {virtual/augmented reality (VR/AR)~\citep{holynski2018fast, kalia2019real}. } 
Among depth cameras, time-of-flight (ToF) technology emerges as the preferred choice for many applications thanks to its compact structure, high precision, and low cost. 
{Recently, the iPad Pro and iPhone Pro/Pro Max from Apple use LiDAR based on direct ToF (dToF) technology, often adopted in fields such as biometrics, photography, games, modeling, virtual reality, and augmented reality.} 
\revise{The ToF-only imaging system is usually preferred when cost efficiency, energy efficiency, fast real-time, or privacy sensitivity are the primary considerations.} 

\revise{While ToF-only cameras might be the preferred choice in scenarios where depth sensing is paramount, some drawbacks still limit their performance in applications such as scene understanding, mixed reality, human-computer interaction, and facial recognition, where accuracy, versatility, and image content cues play a crucial role.}
In particular, the resolution of existing consumer-grade ToF cameras is about 300K pixels, with the LiDAR sensors -- again based on the time of flight principle -- used, for instance, by autonomous vehicles capturing even sparser measurements. 
In contrast, the resolution of RGB cameras used in mobile phones (e.g., Huawei P60) can reach dozens of MegaPixels, i.e., orders of magnitude higher, yet at a much lower cost. 
{One solution is to directly enhance the low-quality (LQ) ToF depth map, i.e., to upsample~\citep{riegler2016atgv} or densify~\citep{uhrig2017sparsity} it, relying on hand-crafted or neural network-learned priors. However, these priors are likely to cause the predicted High-Resolution (HR) depth to contain blurred edges and other undesired artifacts. 
To face these issues, \textit{RGBD cameras} integrate an HR RGB imager and a ToF sensor. }
Then, under the guidance of HR color images, sparse or Low-Resolution (LR) depth maps can be densified or super-solved by exploiting the structural details available from the high-resolution image content. 

{In practical applications, ToF cameras based on different technical routes are deployed in different scenarios, mainly divided into the following three categories: 1) for short-ranging acquisition in applications like face/gesture recognition, typically using an indirect ToF (iToF) camera with a uniform field of illumination, obtaining a dense depth map, yet at a low resolution; 2) for long-range measurements, as in the case of LiDAR sensors emitting laser pulses, which produces a scattered point cloud of the sensed scene: when coupled with a color camera, the sparsity of the depth map obtained by projecting the point cloud on the image plane depends on the resolution of the image, the higher, the sparser -- i.e., containing only $5\%$ of pixels encoding valid depth measurements \citep{Ma2018SelfSupervisedSS} on $\sim0.3$Mpx images;
3) for medium-ranging, as in the case of the ToF cameras used in augmented reality and automated guided vehicle robots, either iToF or dToF can be used.
However, even when iToF is used for short-ranging, depth values around the corners or edges may be missing. }

\begin{figure}[t] 
	\centering
	\renewcommand\tabcolsep{1.5pt} 
    \begin{tabular}{ccc}    
    \multicolumn{3}{c}{\includegraphics[width=0.47\textwidth]{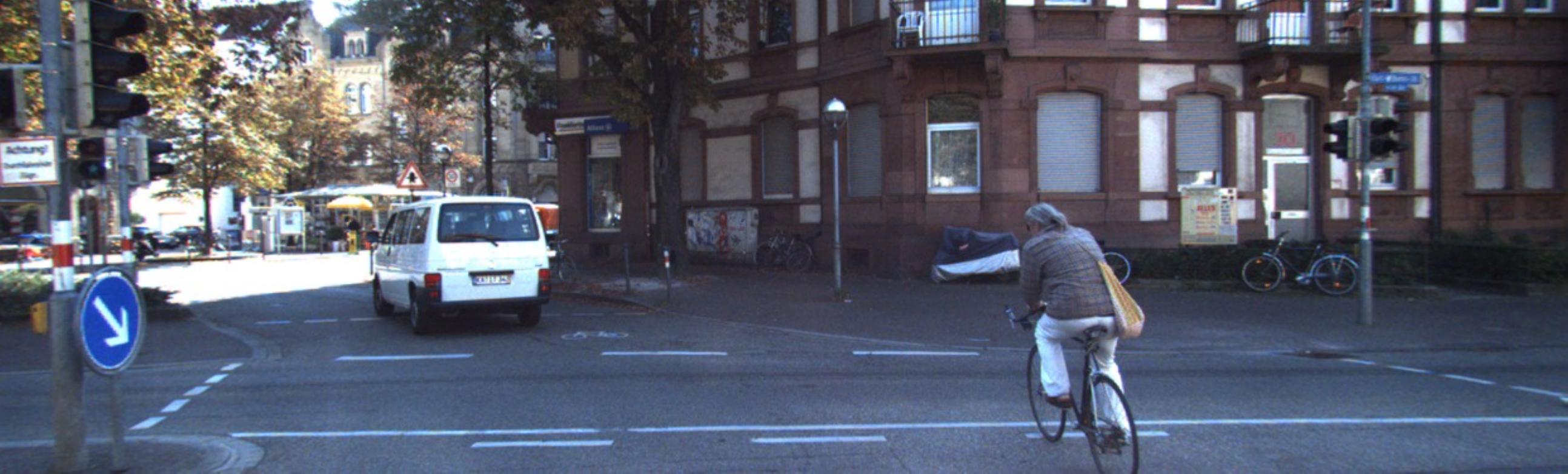}} \\
    \multicolumn{3}{c}{
    \begin{overpic}
     [width=0.47\textwidth]{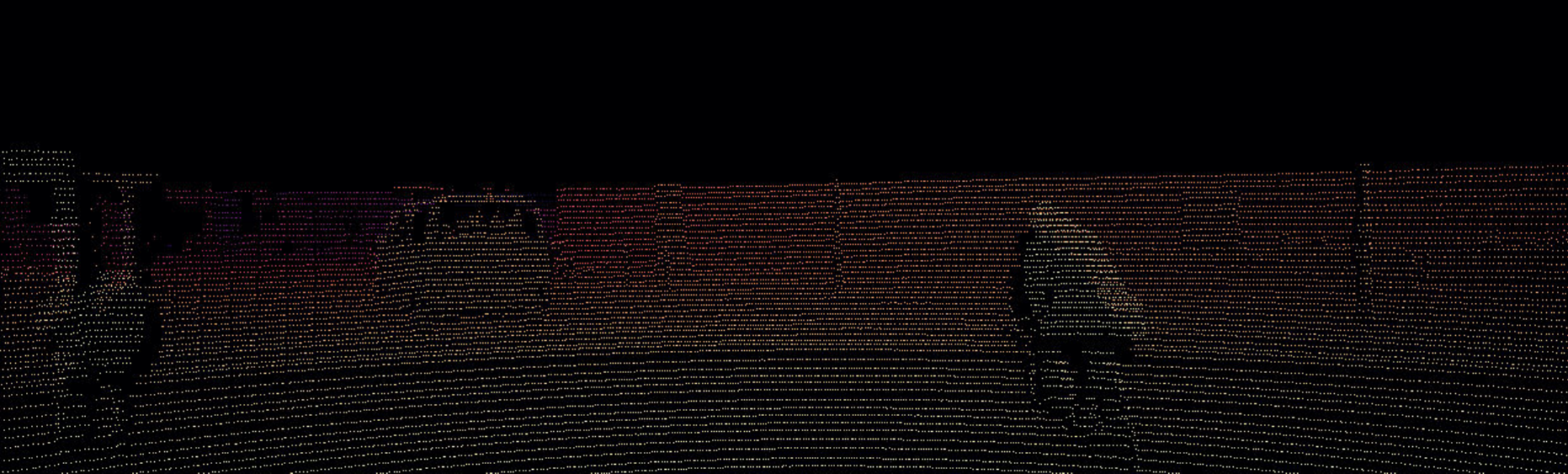}
     \linethickness{1.75pt}
     \put(62,18){\color{red}\line(2,0){12}}
     \put(62,1){\color{red}\line(2,0){12}}
     \put(62,1){\color{red}\line(0,2){17}}
     \put(74,1){\color{red}\line(0,2){17}}
    \end{overpic}} \\
    \multicolumn{3}{c}{375$\times$1242} \\
    \includegraphics[width=0.15\textwidth]{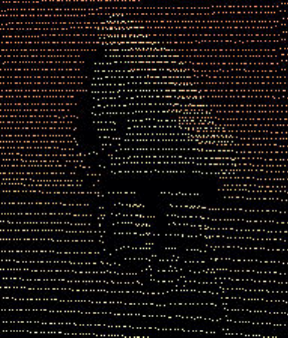} &
    \includegraphics[width=0.15\textwidth]{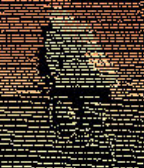} &
    \includegraphics[width=0.15\textwidth]{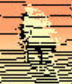} \\
    375$\times$1242 & $\times0.5$ & $\times0.25$ \\
    \includegraphics[width=0.15\textwidth]{crop_x1.00.pdf} &
    \includegraphics[width=0.15\textwidth]{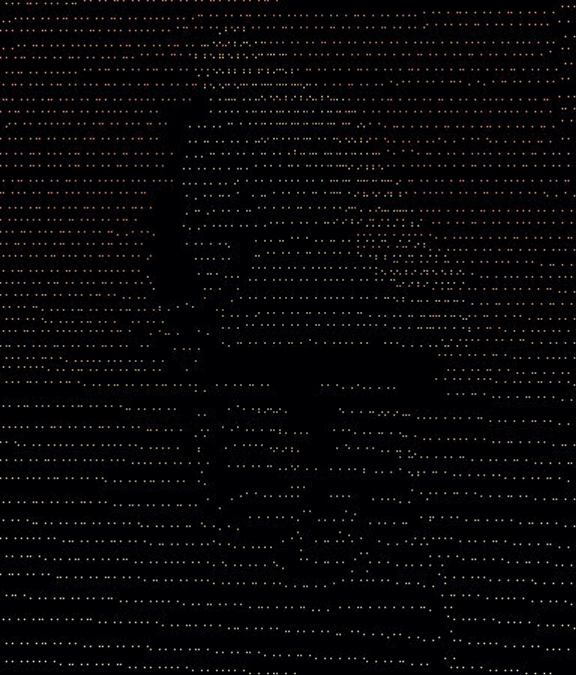} &
    \includegraphics[width=0.15\textwidth]{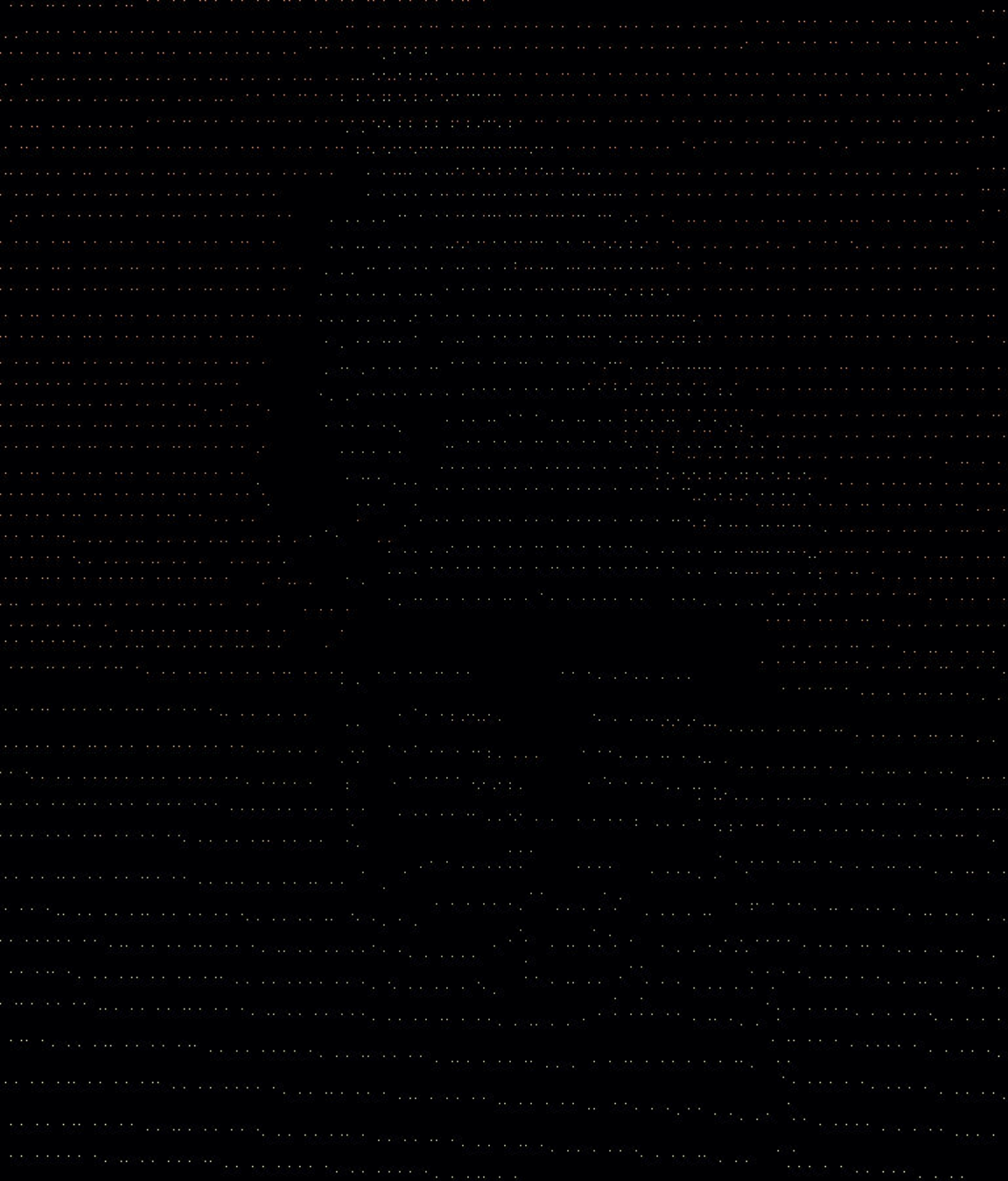} \\
    375$\times$1242 & $\times2$ & $\times4$ \\
    \end{tabular}
    \caption{\revise{An example of LiDAR point cloud from the KITTI dataset reprojected on images at different resolutions.}}
    \label{GDSR_lidar}
\end{figure}

{In the three cases, improving the quality of original depth maps translates into} 1) upsampling the original, spatial resolution through \underline{G}uided \underline{D}epth \underline{S}uper-\underline{R}esolution (\underline{GDSR})~\citep{ferstl2013image,he2021towards}, or 2) densifying the sparse depth map through \underline{G}uided \underline{D}epth \underline{C}ompletion (\underline{GDC})~\citep{uhrig2017sparsity,chodosh2019deep}.
The goal of GDSR is to utilize RGB images to increase pixel density for low-resolution depth and provide richer visual details, while GDC aims to fill in missing depth values in sparse depth maps obtained when using laser pulses, as well as increase the overall depth map resolution -- as the sparsity of the depth map depends on the resolution of the image over which the sensed point cloud is projected (the lower the image resolution, the denser the projected depth map; the higher the image resolution, the sparser the depth map). 
\revise{In Fig. \ref{GDSR_lidar}, we show an example of LiDAR point cloud from the KITTI dataset reprojected at the original resolution used in the dataset (375$\times$1242) as well as both at lower and higher resolution (respectively $\times$0.5, $\times$0.25, $\times$2 and $\times$4). 
We can observe how assuming different camera resolutions produces depth maps whose density changes sensibly, with fewer holes present at lower resolutions and much sparser depth points at higher resolutions. This evidence confirms that the density of a depth map obtained from LiDAR scans is a function of the resolution at which we wish to obtain a depth map and that, to some extent, the completion process has analogies with the super-resolution task.}
\revise{The key difference with GDSR is that the laser spot emitted by LiDAR in GDC is tiny and typically cannot cover the scene corresponding to one pixel in the sensor. }
\revise{Therefore, GDC usually requires perceiving a larger neighborhood (e.g., $7\times 7$) than GDSR in shallow feature extraction of the input to estimate depth values in areas where the information is missing.}
Generally, RGB guided ToF imaging systems involve one of the two sub-tasks, depending on the use case and the depth sensing technologies. Consequently, different solutions must be deployed depending on the specific sub-task.
\revise{Nonetheless, both tasks aim to yield high-quality depth maps with the guidance of RGB information, that is, transfer high-frequency structural information and fine-grained features from HR color images to LR/sparse depth maps while avoiding texture-copying artifacts. 
Furthermore, it is worth noting that the model design of the two tasks exhibits intriguing similarities, which inspired us to analyze them together as one topic for investigation, and their specific classification will be elaborated upon in Sec. \ref{taxo}.}

\begin{figure*}[t]
   \centering
    \begin{tabular}{c}
		\includegraphics[width=0.97\linewidth]{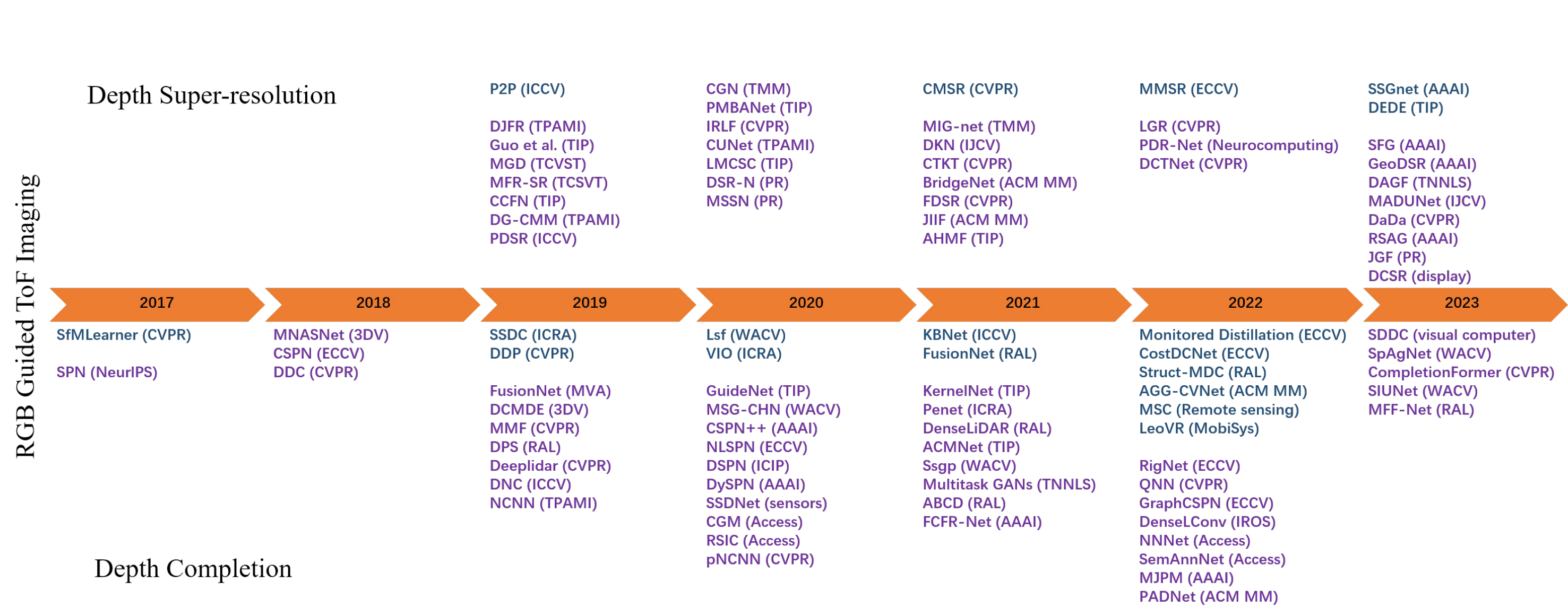} 
    \end{tabular}
    \caption{Timeline showing the advances in RGB guided ToF Imaging. Methods are marked in blue or purple, depending on the learning paradigm being unsupervised or supervised.}
    \label{method_timeline}
\end{figure*}

At first, traditional methods extract meaningful features from color images through guided filtering-based~\citep{petschnigg2004digital,he2012guided,liu2012guided} or optimization-based~\citep{diebel2005application,ferstl2013image,lu2014depth} algorithms.
These hand-crafted approaches rely on the assumption that depth discontinuities statistically co-occur in correspondence with RGB image edges.
In real-world cases, however, this prior may fail to extract informative features and introduce artifacts. For instance, in the presence of rich texture, these methods convey essential structural details but also transfer the appearance of RGB images to depth maps as a side effect.
Therefore, this task remains challenging as it requires properly distinguishing between high-frequency image information corresponding to discontinuities in the depth map and those not.
Over the past decades, solutions for enhancing RGB guided ToF imaging have shifted from model-driven to data-driven paradigms~\citep{deng2019coupled,wong2021learning}, in particular thanks to deep learning methods~\citep{cheng2020cspn++,zuo2020frequency,zhao2022discrete} that have rapidly advanced the field.
Even with a simple network structure, such as a few stacked convolutional layers~\citep{uhrig2017sparsity} or the vanilla encoder-decoder~\citep{li2016deep} architecture, it is straightforward to recover HR/dense depth maps with much higher accuracy compared to what obtained through hand-crafted methods.
Consequently, RGB guided ToF imaging via deep learning received increasing research interest from academia and industry, motivating us to survey recent progress in this field.

Specifically, given the substantial development of deep-learning solutions in the last decade, this survey aims to provide a complete overview of solutions designed for enhancing RGB guided ToF imaging in both use cases. 
Although we are aware of previous review papers on guided depth super-resolution \citep{zhong2023guided} and depth completion \citep{survey_completion_1,survey_completion_2} treated as standalone tasks, we highlight the lack of interconnection between the two, despite the shared insights they bring to the community of researchers working with RGB guided ToF sensors, which we aim to envelop in a comprehensive survey. 
Purposely, we will review the literature concerning deep learning frameworks developed to enhance the depth maps perceived from this family of devices. We will study and classify them according to several aspects, such as the framework design, the learning paradigm, and the objective functions minimized during training.
Moreover, we introduce the reader to the standard datasets used as benchmarks for evaluating existing methodologies, reporting the performance of state-of-the-art methods as a reference.

We outline our major contributions as follows:
\begin{itemize}
    \item {To the best of our knowledge, we are the first to provide an in-depth investigation for RGB guided ToF imaging through deep learning, including guided depth super-resolution and guided depth completion as the main tasks to be dealt with according to the specific use case. 
    }
    \item {In each subtask, this paper reviews recent deep learning-based methods under several aspects to illustrate the motivations behind their design, contributions, and performance.}
    \item {We discuss the challenges and future trends of RGB guided ToF imaging.}
\end{itemize}

The rest of this survey is structured as follows. In Section~\ref{fund}, we describe preliminaries for RGB guided ToF imaging. Section.~\ref{dsr} and \ref{dcp} present deep learning-based methods for GDSR and GDC, respectively. In Section~\ref{datasets}, we introduce the datasets and loss functions. Section~\ref{exp} provides comparison results of some recent methods. Finally, we discuss future trends and challenges in Section~\ref{future}. Fig. \ref{method_timeline} draws a timeline and reports for each year from 2017 to 2023, any method we will review in this paper, and the publication venue. 

\section{Preliminaries and Taxonomy} \label{fund}
In this section, we first introduce the foundation of RGB guided ToF imaging, then we present GDSR and GDC as the two main research trends aimed at enhancing ToF depth maps and the commonly used objective functions in both.

\subsection{Principles of ToF Imaging}
\begin{figure}[t]
   \centering
    \begin{tabular}{c}
		\includegraphics[width=0.93\linewidth]{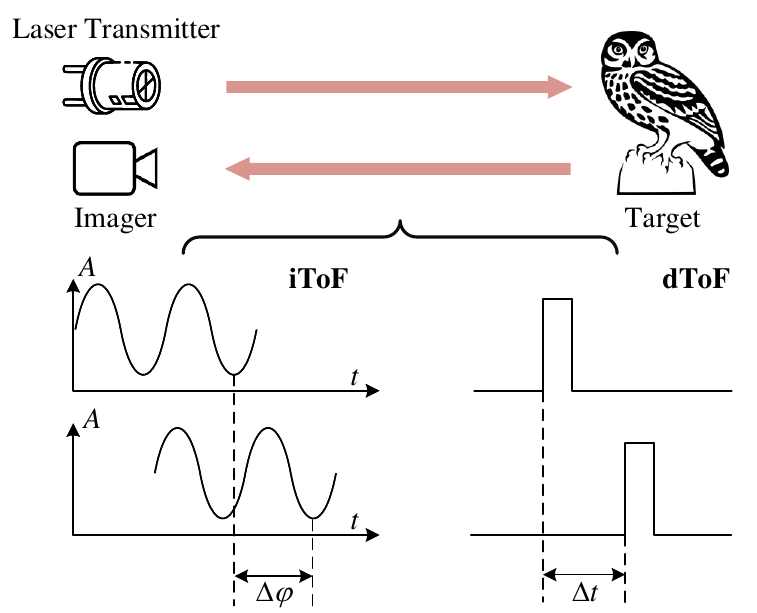} 
    \end{tabular}
    \caption{{\textbf{iToF and dToF working principles.} The former computes distances based on the phase shift between emitted and reflected light; the latter measures the time required for a laser light to bounce over an object.}}
    \label{fig_tof_principle}
\end{figure}

Here, we describe the imaging principles of iToF and dToF, respectively. In Fig.~\ref{fig_tof_principle}, we provide a schematic diagram of the principles commonly used by iToF and dToF. 

\subsubsection{Indirect ToF}
Indirect ToF (iToF) determines distance by capturing reflected light and calculating the phase delay between emitted and reflected light. Currently, most of the off-the-shelf ToF cameras emit signals $s(t)$ using sinusoidal amplitude modulation:
\begin{equation}
	s(t)=s_{1}\cos (\omega t)+s_0
	\label{eq1}
\end{equation}

\begin{table*}[htbp]	\scriptsize
	\centering
        \renewcommand\arraystretch{1.4}
	\renewcommand\tabcolsep{10pt}
	\caption{\revise{Characteristics of different types of LiDAR.}}
	\begin{tabular}{@{}lccc@{}} 
		\toprule
		\textbf{Type} & \textbf{Technology} & \textbf{Advantages} & \textbf{Disadvantages}   \\
		\midrule
            Mechanical &  Rotating mirrors or prisms &  High accuracy, long range  & Large, expensive  \\ \hline
            MEMS & Micromirrors & Small, cheap  &  Lower FoV, range   \\ \hline
            Flash & Arrays of lasers firing simultaneously  &  Fast, high resolution  &  Short range  \\  \hline
            OPA & Optical phased arrays & Small, cheap, versatile  & Still in development   \\ 
		\bottomrule
	\end{tabular}
	\label{tab_LiDAR}
\end{table*}

Since the baseline between the ToF transmitter and the receiver is so small (usually only a few millimeters), they can generally be considered coaxial. Assuming that the target is static, and the signal is only reflected once before being received by the sensor, the reflected signal of a single pixel measured at a specific fixed frequency can be expressed as a function of time $t$:
\begin{equation}
	r(t)=\frac{\rho}{c^{2}\tau_{0}^{2}}[s_{1}\cos (\omega t-2\omega \tau_{0})+s_{0}]+e_{0}
\end{equation}
where $\rho$ is the surface reflectivity of the target, $\omega$ denotes angular frequency, $\tau_{0}$ is the delay of signal reflection from the target back to a pixel at the speed of light $c\approx 3\times 10^8$m/s, and $e_{0}$ is the offset caused by ambient light. According to \cite{heide2015doppler}, Eq. \ref{eq1} can be rewritten as:
\begin{gather}
	A_1=\frac{\rho s_{1}}{c^{2}\tau_{0}^{2}}, \quad A_{0}=\frac{\rho s_{0}}{c^{2}\tau_{0}^{2}}+e_{0}   \\
	r(t)=A_1\cos (\omega t-2\omega \tau_{0})+A_{0}
	\label{eq2}
\end{gather}
The received signal remains sinusoidal, but the phase, which contains scene depth information, is changed. After a high-frequency periodic signal $B_{1}\cos(\omega t-\phi)$ modulates the incident signal, the modulated signal can be expressed as:
\begin{equation}
	\label{eq3}
	\begin{aligned}
		\Tilde{I}_{\phi ,\omega}(t)=&r(t)B_{1}\cos (\omega t-\phi)    \\
		=&\frac{A_{1}B_{1}\cos(\phi-2\omega \tau_{0})}{2}+ \\ &\frac{A_{1}B_{1}\cos(2\omega t+\phi+2\omega \tau_{0})}{2}+\\ &A_{0}B_{1}\cos(\omega t-\phi)
	\end{aligned}
\end{equation}
Considering the quantum efficiency of ToF sensors, the sensor exposure time $T$ set to capture a sufficient number of photons is usually much larger than $\pi c/\omega$. During the exposure, the measurement is equivalent to integrating over the time range $T$, so the period terms in Eq. \ref{eq3} all vanish in the calculation, and the result can be computed as:
\begin{align}
    I_{\phi,\omega}&=\int_{-T/2}^{T/2}\Tilde{i}_{\phi,\omega}(t)dt \nonumber \\
    &\approx A_{1}B_{1}\cos(\phi-2\omega \tau_{0})=A_{1}g_{\phi,\omega}
	\label{eq4}
\end{align}
where $A_1$ is the scene response that represents the intensity of the encoded corresponding pixel at time $\tau_{0}$ after the laser transmitter sends a light pulse, $g_{\phi,\omega}=B_1\cos(\phi-2\omega \tau_{0})$ is the camera function, and $I_{\phi,\omega}$ is the raw measurement with correlation at phase $\phi$, angular frequency $\omega$, i.e. ToF raw data. If a single modulation frequency $f_m$ is used to collect multiple ($K\ge 2$) raw measurements in one cycle, the distance between the scene and camera can be calculated as:
\begin{equation}
		d=\frac{c}{2\omega}\arctan (\frac{\sin{\boldmath{\phi}}\cdot \boldmath{I}_{\phi,\omega}}{\cos{\boldmath{\phi}}\cdot \boldmath{I}_{\phi,\omega}})   \\
	\label{eq5}
\end{equation}
In practice, if the modulation frequency $f_m=\frac{\omega}{2\pi}$ is fixed, ToF cameras usually adopt the four-phase method to sample ToF raw data, i.e., sampling four times in one cycle, with a phase step of $90^\circ$. For ease of reading, the four raw data can be abbreviated as $I_{i}=I_{\frac{\pi}{2\omega}i}, i=0,1,2,3$.
\begin{gather}
	A=\frac{\sqrt{(I_{3}-I_{1})^{2}+(I_{0}-I_{2})^{2}}}{2}     \label{eq_amp}
	\\
	B=\frac{I_{0}+I_{1}+I_{2}+I_{3}}{4}     \label{eq_offset}
	\\
	\phi=\arctan{\frac{I_{3}-I_{1}}{I_{0}-I_{2}}}       \label{eq_phi}
	\\
	d=\frac{c}{2\pi}(\frac{\phi}{2f}+N)          \label{eq_depth}
\end{gather}
where $N$ is an integer that represents the number of phase wraps. Based on the above principle, some works~\citep{su2018deep,chugunov2021mask,gutierrez2021itof2dtof,li2022fisher} directly process the raw signal of the iToF, thus improving the depth quality. 

At present, most iToF for consumer use sensors based on complementary metal-oxide semiconductor (CMOS) or charge-coupled device (CCD), with a resolution of $320\times 240$ (QVGA) or $640\times 480$ (VGA) pixels. However, the resolution is still significantly lower than that of an RGB image using the same technology. 

\begin{figure*}[t]	
	\centering	
	\subfloat[Principle of mechanical LiDAR]{	
		\centering	
		\includegraphics[height=1.7in]{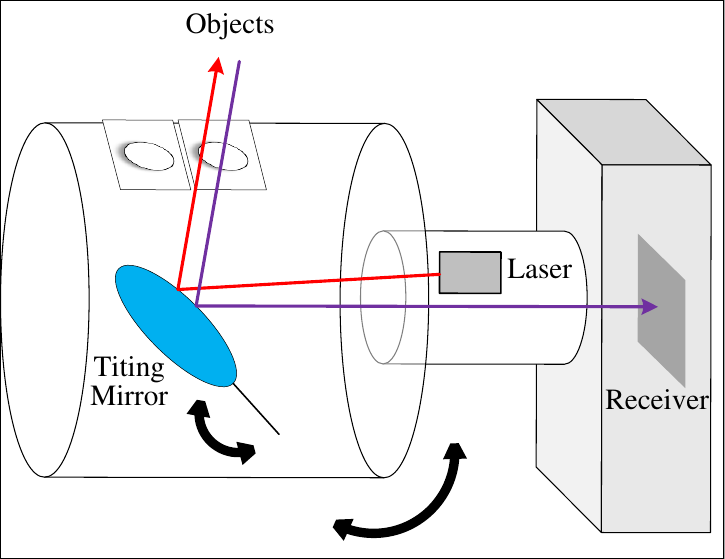}}	
	\hspace{2mm}
	\subfloat[Principle of MEMS LiDAR]{	
		\centering	
		\includegraphics[height=1.5in]{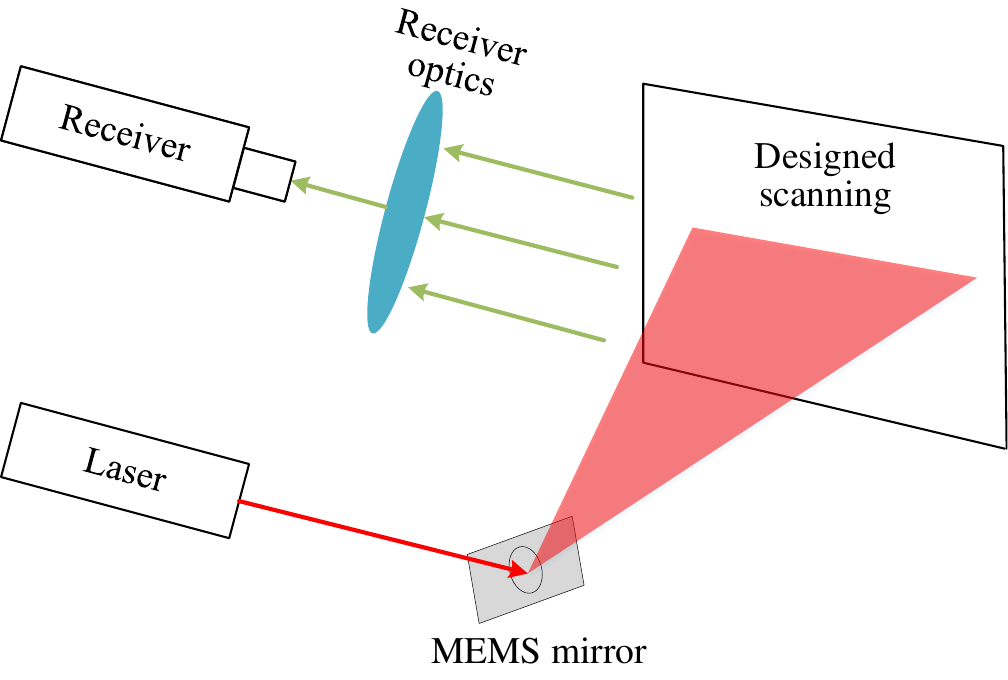}}
 
	\subfloat[Principle of flash LiDAR]{	
		\centering	
		\includegraphics[height=1.5in]{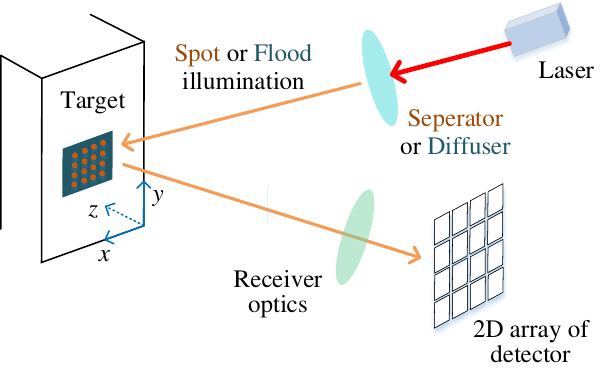}}
        \hspace{2mm}
	\subfloat[Principle of OPA LiDAR]{	
		\centering	
		\includegraphics[height=1.5in]{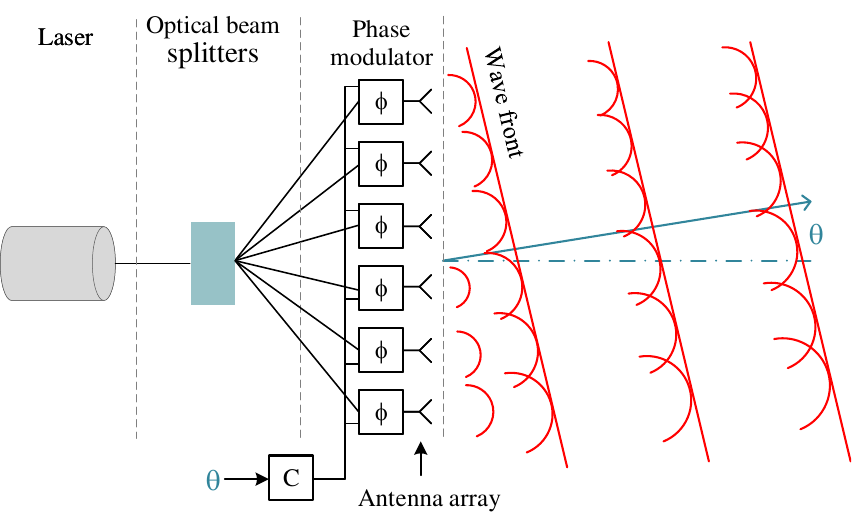}}
	\caption{\revise{Schematic diagram of LiDAR systems with different imaging principles.}}
	\label{fig_lidars} 
\end{figure*}

\subsubsection{Direct ToF}
DToF is a method that measures the time it takes for laser light to be fired directly at an object and returned to the sensor. The moment the laser is fired, the electronic clock is activated. Then, the pulse bounces off the target and is partially picked up by the photodetector. By measuring the time lapse $\Delta T$, the distance $d$ between the outgoing light pulse and the reflected light pulse is calculated with
\begin{equation}
    d=\frac{c\Delta T}{2}   \\
	\label{eq_dtof}
\end{equation}
To achieve higher sensitivity and lower power consumption, dToF usually adopts single photon avalanche diodes (SPADs) instead of Avalanche Photodiodes (APDs) as photodetectors. Due to limitations in the size of the chip and its auxiliary circuits, the pixel count of SPAD arrays is much lower than that of iToF. 

\textbf{LiDAR}, an acronym for Light Detection and Ranging, despite being based on the same working principle as dToF, is frequently listed as a distinct depth sensing technology due to some of its peculiar design and setup. For instance, conventional LiDAR continuously scans the environment utilizing mechanical components (e.g., spinning mirrors). 
Another type is MEMS LiDAR, which uses micromirrors controlled by micro-electromechanical systems (MEMS) to scan the environment. This type of LiDAR is smaller and cheaper than mechanical LiDARs but has lower FoV and range. 
As a still developing technology, OPA LiDARs use optical phased arrays (OPAs) to steer the laser beam electronically, and they have the potential to be small, cheap, and versatile.
Different from the techniques mentioned above, global shutter flash LiDARs, also referred to as flash LiDARs or ToF LiDARs, use arrays of lasers firing simultaneously to create a 3D image of the surrounding environment, just like iToF, thereby improving frame rate and imaging efficiency for applications such as autonomous driving.
This technology enables the integration of dToF into consumer electronic devices. We provide a comparison summarizing the different types of LiDAR in Table.~\ref{tab_LiDAR} and the schematic diagram of each method in Fig. \ref{fig_lidars}.

In essence, LiDAR still employs the dToF principle, and the two terms of dToF and LiDAR are even interchangeable in many scenarios (e.g., Apple refers to the dToF sensor in iPad Pro and iPhone 12 Pro as a LiDAR scanner).
However, LiDAR technology is rapidly evolving, and new state-of-the-art LiDARs are constantly being developed, thus potentially having more applications in various scenarios. For instance, the InnovizOne MEMS LiDAR, manufactured by EDOM technology, achieves an angular resolution of up to $0.1^{\circ} \times 0.1^{\circ}$, producing much denser point clouds than before.

\begin{table*}[htbp]	\scriptsize
	\centering
        \renewcommand\arraystretch{1.4}
	\renewcommand\tabcolsep{15pt}
	\caption{{Characteristics of iToF and dToF.}}
	\begin{tabular}{@{}lcc@{}} 
		\toprule
		\textbf{Parameter} & \textbf{iToF} & \textbf{dToF}    \\
		\midrule
            Principle & \tabincell{c}{Using phase shift between emitted \\ and reflected light to determine distance.}  & \tabincell{c}{Using stop watch method \\ to calculate time lapse.} \\ \hline
            Detector type & \tabincell{c}{PMDs: $6\sim 100 \mu m$ pixel size} & SPADs / APDs \\ \hline
            Depth calculation & In-pixel calculation & Histogram analysis \\ \hline
		Performance & Long integration time & Fast acquisition \\ \hline
            Range & Short-medium  & Longer  \\ \hline
		  Range ambiguity & Yes & No \\ \hline
		  Pixel count & Large & Smaller \\ \hline
		Accuracy & Linearly related to distance & Higher \\ \hline
            Power consumption & Low & Higher \\ \hline
            Portability & Generally compact & Traditionally larger \\ \hline
            Cost & Medium & Higher \\ 
		\bottomrule
	\end{tabular}
	\label{tab_iToF_dToF}
\end{table*}

\subsubsection{Comparative Analysis of dToF and iToF}
Here, we provide their characteristics in Table \ref{tab_iToF_dToF}. Besides, we conduct a comparative analysis on the following aspects.

\textbf{Accuracy.} iToF cameras typically provide accurate distance measurements within a specific range. Its measurement error is theoretically positively correlated with the target distance. 
Currently, the accuracy of mass-produced iToF cameras can be controlled within $1\%$, e.g., Helios2 ToF 3D Camera with Sony’s IMX556 DepthSense ToF has an accuracy of less than 5mm and a precision of less than 2mm at 1m camera distance. At the same time, dToF (LiDAR) can achieve millimeter-level accuracy even over longer distances. Therefore. for short-range applications, both iToF and dToF can achieve comparable accuracy, although they may fall short of structured light cameras. However, a key advantage of LiDAR is that its accuracy substantially remains constant regardless of distance, as discussed in Sec 2.1.3. This attribute positions LiDAR measurements as reliably accurate in long-range applications, distinguishing them from other technologies.

\textbf{Range and Power consumption.} iToF cameras are generally suitable for short to medium-range applications. The effective range may vary but typically within a few to tens of meters, depending on the specific design and implementation. Because of these characteristics, ToF cameras are often designed to be power-efficient and, therefore, suitable for consumer-grade devices. In contrast, LiDAR systems can achieve longer ranges, making them suitable for applications like autonomous vehicles and long-range mapping. Consequently, they are commonly employed in scenarios where power consumption is less critical.

\textbf{Portability.} iToF cameras are inherently compact and lightweight, making them suitable for integration into portable devices such as smartphones, tablets, and wearable gadgets. Meanwhile, LiDAR, traditionally larger, is undergoing miniaturization trends, especially with the development of solid-state LiDAR, making it increasingly applicable in portable and handheld devices. 

\textbf{Applications.} Due to the properties peculiar to each, the two families of sensors are usually involved in different downstream applications. Specifically, iToF sensors are the best fit for AR/VR experiences, which mainly focus on narrow, close-range environments or, in general, for any applications running in a constrained space (e.g., bin picking with a robotic arm).
On the contrary, dToF is preferable in contexts requiring long-range perception, e.g., the automotive one with ADAS or autonomous driving systems. 
It is worth noting that there is an overlap in the operating ranges of iToF and dToF, i.e., both can be used for applications such as AR/VR (e.g., Kinect v2 with iToF and iPad Pro with dToF) when performing medium-range sensing.

Though both imaging systems currently excel in their respective application domains, advancements in technology are steadily bridging the gap between their performance metrics, leading to a growing overlap in their practical uses.

\subsection{Problem Formulation}        
{Although ToF cameras have many advantages, they also bring some inherent limitations, such as multipath interference~\cite{gupta2015phasor,gutierrez2021itof2dtof}, flying pixels~\cite{qiao2020valid} and wiggling errors~\cite{hussmann2013modulation}. 
Hand-crafted models can deal with some issues, e.g., depth correction, but others require additional cues, such as those available in RGB images. }

The goal of RGB guided ToF imaging is to recover a high-quality depth map $Z_{hq}$ from the acquired low-quality depth map $Z_{lq}: \Omega_{Z} \subset \Omega \mapsto \mathbb{R}_{+}$ and high-resolution color image $I: \Omega \subset \mathbb{R}^{2} \mapsto \mathbb{R}^{3}_{+}$. Depending on the setup of ranging systems, low-quality depth maps can be either dense, yet at low resolution, or sparse. Besides, we assume all LQ depth maps and the corresponding color images are correctly aligned. 
This is possible if the RGB-D system is calibrated, i.e., the relative pose between the two color and depth cameras is known.
If a deep neural network $\Phi_{\gamma}$ with parameters $\gamma$ is deployed for RGB guided ToF imaging, the task can be modeled as
\begin{equation}
    \Hat{Z}_{hq}=\Phi_{\gamma}(Z_{lq}, I)
\end{equation}
where $\Hat{Z}_{hq}$ is the prediction of latent HR depth map. 

The network parameters $\gamma$ are updated during training of $\Phi$, by solving the following optimization problem:
\begin{equation}
    \Hat{\gamma}=\mathop{\arg\min}\limits_{\gamma} \mathcal{L}(Z_{hq}, \Hat{Z}_{hq})
\end{equation}
where $\mathcal{L}$ is an objective function, usually minimizing the distance between the prediction and the ground truth depth.

\subsection{Evaluation metrics}
As for most computer vision tasks, for RGB guided ToF imaging, it is necessary to employ appropriate evaluation metrics. 
The most commonly used measures for both the two sub-tasks, i.e., GDSR and GDC, are root mean squared error (RMSE) and mean absolute error (MAE):
\begin{gather}
    \mathrm{RMSE}(mm)=\sqrt{\frac{1}{N}\sum_{p \in N}(Z_{hq}^{p}-\Hat{Z}_{hq}^{p})^2}           \\N
    \mathrm{MAE}(mm)=\frac{1}{N}\sum_{p\in N}|Z_{hq}^{p}-\Hat{Z}_{hq}^{p}|
\end{gather}
with $p$ being a single pixel in the depth map.

\textbf{Metrics for GDSR.} In addition to the above metrics, mean squared error (MSE) is frequently utilized, which has a similar role to RMSE:
\begin{equation}
    \mathrm{MSE}=\frac{1}{N}\sum_{p\in N}|Z_{hq}^{p}-\Hat{Z}_{hq}^{p}|^2
\end{equation}
Furthermore, with a short-range ToF sensor, GDSR focuses on recovering depth maps with desirable details. As a result, peak signal-to-noise ratio (PNSR) and structural similarity index (SSIM) are also sometimes used to assess the quality of depth maps, which are defined as:
\begin{gather}
    \mathrm{PSNR}=10log_{10}(\frac{Z_{max}^2}{\mathrm{MSE}})    \\
    \mathrm{SSIM}(p,q)=\frac{(2\mu_{p}\mu_{q}+C_{1})(2\sigma_{xy}+C_{2})}{(\mu_{p}^{2}+\mu_{q}^{2}+C_{1})(\sigma_{p}^{2}+\sigma_{q}^{2}+C_{2})}
    \label{ssim}
\end{gather}
with $Z_{max}$ being the maximum depth value; $q$ is a neighbor of pixel $p$; $\mu_{p}$ and $\mu_{q}$ denote the mean values of HQ depth maps and the corresponding ground truth, respectively; $\sigma_{p}^{2}$ and $\sigma_{q}^{2}$ are the variance; $C_1$ and $C_2$ are constants used to maintain the stability of the division. The smaller the pixel value difference between the two depth maps, the higher the PSNR. 

\textbf{Metrics for GDC.} Sparse depth measurements are often captured from a long-range ToF LiDAR, so there are several task-specific metrics for GDC, including RMSE of the inverse depth (iRMSE), MAE of the inverse depth (iMAE), which are defined by
\begin{gather}
    \mathrm{iRMSE}(\frac{1}{km})=\sqrt{\frac{1}{n}\sum_{n\in N}(Z_{hq}^{p}-\Hat{Z}_{hq}^{p})^2} \\
    \mathrm{iMAE}(\frac{1}{km})=\frac{1}{N}\sum_{p\in N}|\frac{1}{Z_{hq}^{p}}-\frac{1}{\Hat{Z}_{hq}^{p}}|
\end{gather}

When the evaluation of deep models is performed on indoor datasets, e.g., NYU-v2~\citep{silberman2012indoor}, mean absolute relative error (REL) and thresholded accuracy are more popular
\begin{gather}
    \mathrm{REL}(mm)=\frac{1}{N}\sum_{n\in N}\frac{|Z_{hq}^{p}-\Hat{Z}_{hq}^{p}|}{Z_{hq}^{p}} \\
    \sigma=\max(\frac{Z_{hq}^{p}}{\Hat{Z}_{hq}^{p}}, \frac{\Hat{Z}_{hq}^{p}}{Z_{hq}^{p}})<th
\end{gather}
with $th$ being a given threshold. 

\revise{\subsection{Taxonomy}
\label{taxo}

\begin{figure}[t]
   \centering
    \begin{tabular}{c}
		\includegraphics[width=0.98\linewidth]{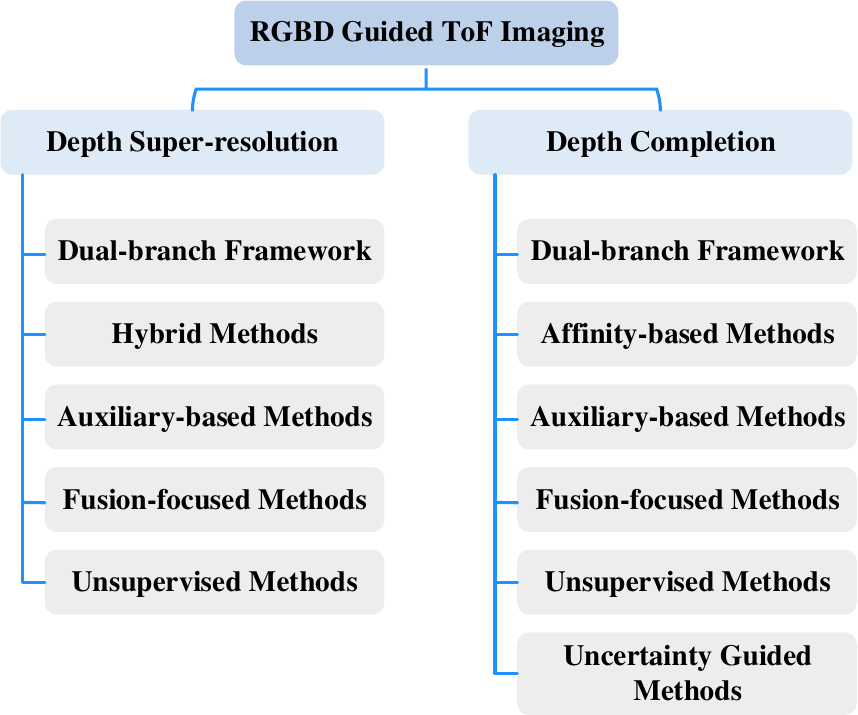} 
    \end{tabular}
    \caption{\revise{Taxonomy of RGB guided ToF imaging techniques. We identify 5 and 6 main categories for GDSR and GDC methods.}}
    \label{method_overview}
\end{figure}

Figure \ref{method_overview} anticipates the proposed taxonomy for methods we will survey in this paper. 
On the one hand, from the figure, we can already highlight some common trends in GDSR and GDC tasks. Specifically, we can identify the category of \textit{Dual-branch Frameworks} in both, i.e., those deploying two main branches for processing RGB and depth cues separately, as well as the category of \textit{Auxiliary-based Methods} -- those exploiting multi-task learning to improve the accuracy of the predicted depth map -- \textit{Fusion-focused Methods} which mainly study the fusion strategy for combining color and depth features, and the \textit{Unsupervised Methods} not requiring ground-truth labels at training time. On the other hand, we identify some trends that are peculiar to one of the two tasks: specifically, for GDSR, we have \textit{Hybrid Methods}, focusing on obtaining explainable models to run the super-resolution process; for GDC, \textit{Affinity-based Methods} developed are characterized by the used of Spatial Propagation Networks (SPNs) which learn affinity matrices for propagating depth across neighboring pixels, and \textit{Uncertainty-guided Methods} explicitly involve uncertainty modeling to improve the results.}
Each category will be introduced and discussed in detail in Sections~\ref{dsr} and \ref{dcp}, for GDSR and GDC, respectively.
\section{Guided Depth Super-Resolution} \label{dsr}

In this section, we present and discuss recent deep learning-based approaches. Initially, researchers exploited general architectures {with two branches} to extract features from RGB and depth separately, which are fused in a naive manner, i.e., by means of concatenation or summation. Numerous novel architectures and fusion schemes have been proposed to boost network accuracy. From the perspective of data requirements, depth super-resolution methods are divided into supervised and unsupervised learning. The difference between the two paradigms is that the former requires labeled input data, while the latter does not. For what concerns supervised learning, we classify existing works into three categories, depending on the main contribution they bring to the literature, in terms of ({\romannumeral 1}) Dual-branch framework; ({\romannumeral 2}) optimization-inspired architecture; ({\romannumeral 3}) auxiliary learning; ({\romannumeral 4}) multi-modal fusion scheme. For unsupervised learning, the emphasis is on the fusion strategy and loss functions. These five categories are listed in the top branch of Fig. \ref{method_overview}.

Supervised methods are supplemented by labeled data so that the task can be viewed as a regression problem. Thus, deep models aim to learn depth predictions as close as possible to the ground truth depth labels.

\subsection{Dual-branch Framework} 
\begin{figure}[t]
   \centering
    \begin{tabular}{c}
		\includegraphics[width=0.95\linewidth]{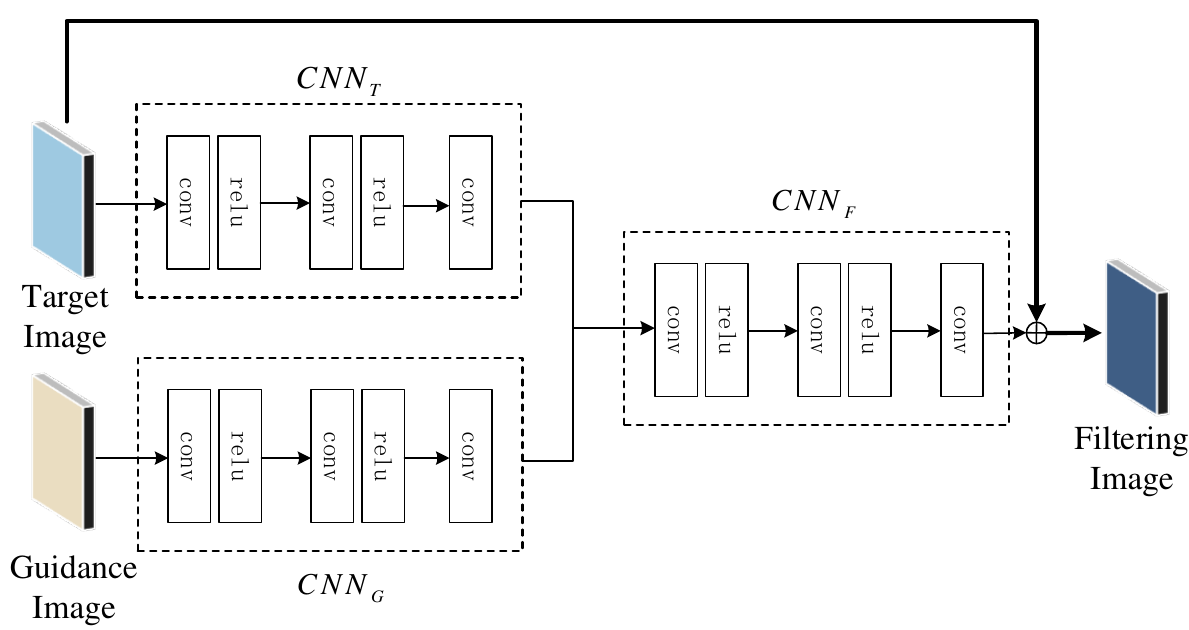} 
    \end{tabular}
    \caption{DJFR architecture~\citep{li2019joint} {belonging to dual-branch framework.} It employs a typical dual-branch auto-encoder framework, where the guide and source features are extracted from the two sub-networks in the encoder.}
    \label{fig:sr_li2019joint}
\end{figure}

Since the depth sensor captures the same scene as the RGB camera, the obtained two images are geometrically similar and complementary. Therefore, how to extract meaningful features from the two modalities through designed algorithms is the key to depth super-resolution. Intuitively, \cite{li2019joint} design a {deep joint filter (DJFR)} which contains two sub-networks for extracting depth and RGB features, respectively. After concatenating the two streams, the fused information is decoded by another sub-network and outputs the predicted depth, as seen in Fig.~\ref{fig:sr_li2019joint}.

Multi-scale information is crucial for many low-level vision tasks, including GDSR, which can recover images from different scales through rich hierarchical features. {The multi-scale guided convolutional network (MSG-Net)}~\citep{hui2016depth} is the first work to super-solve depth maps using multi-scale guidance, which progressively enhances depth details from high- to low-level. Inspired by this idea, numerous methods utilizing multi-scale guidance have been proposed and attained significantly improved performance. For instance, a novel {deep network for depth map super-resolution (DepthSR)} proposed by~\cite{guo2018hierarchical} leverages a residual U-Net architecture~\citep{ronneberger2015u}, where hierarchical guidance is introduced at each scale, to recover HR depth maps. \cite{zuo2019depth} design a multi-scale architecture with a guide similar to MSG-Net~\citep{hui2016depth} where dense layers~\citep{huang2017densely} are employed to revisit the features from all higher levels at a given scale. Also, guided by RGB images, {the multi-scale fusion residual network for GDSR (MFR-SR)} proposed by \cite{zuo2019multi} progressively upsamples depth maps in multiple scales with global and local residual learning. \cite{li2020depth} employ a multi-scale strategy with the proposed symmetric unit (SU) as the basic component of the network. With SU, it can effectively deal with textureless and edge features in depth images. Unlike previous works, \cite{zuo2021mig} propose a two-stream multi-scale network, MIG-net, including three branches, i.e., intensity, depth, and gradient branches.
At each scale, the depth and gradient features are iteratively refined by the guide. 
\cite{zhong2021high} introduce an {attention-based hierarchical multi-modal fusion network (AHMF)} where a bi-directional hierarchical feature collaboration module is proposed to make full use of multi-scale features. In this module, features from different levels are aggregated so that low and high level spatial information can collaborate to improve each other.
\cite{wang2023joint} present a novel method that exploits pyramid structure to capture multi-scale features so that HR depth maps can be progressively recovered.
\cite{yuan2023structure} propose a structure flow-guided framework for GDSR, whose core is to learn a structure flow map to guide structure representation transfer. In this method, a flow-enhance pyramid edge attention network is introduced to learn multi-scale features for sharp edge reconstruction.

\begin{figure}[t]
   \centering
    \begin{tabular}{c}
		\includegraphics[width=0.95\linewidth]{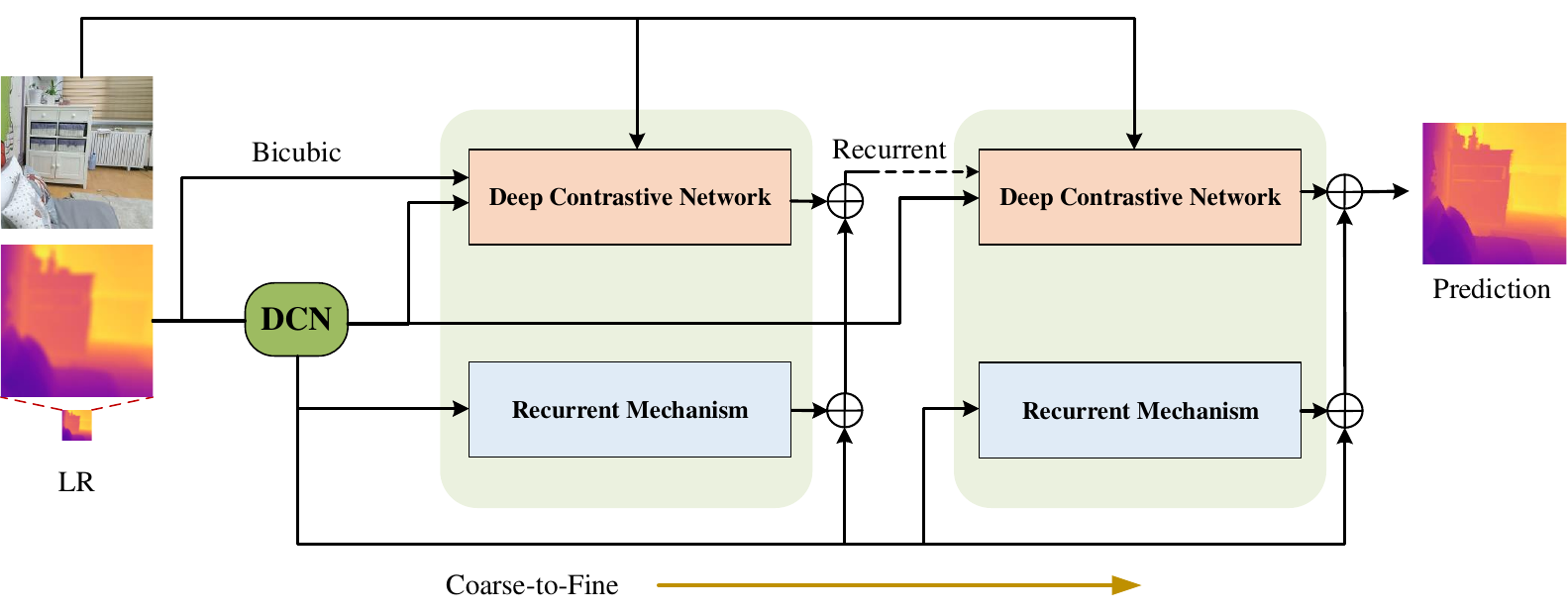} 
    \end{tabular}
    \caption{RSAG architecture~\citep{yuan2023recurrent} {belonging to dual-branch framework.} It progressively refines the depth using a coarse-to-fine strategy.}
    \label{fig:sr_RSAG}
\end{figure}

Using a coarse-to-fine scheme is another common solution for GDSR, consisting of building a multi-stage network to enhance depth maps progressively. The definitions of the multi-scale strategy and the coarse-to-fine scheme may overlap in some approaches. Thus, we distinguish them according to the focus of the works. \cite{wen2018deep} propose a coarse-to-fine neural network with two stages. In the coarse stage, the model learns large filter kernels to obtain less accurate results, further refined by smaller filter kernels in the fine stage. In \cite{zuo2020frequency}, the filtering and refinement of depth-guided intensity features and intensity-guided depth features are iteratively conducted in multiple stages through the proposed depth-guided affine transformation. The network can reduce artifacts produced by distribution gaps between depth maps and the corresponding guide. \cite{ye2020pmbanet} introduce a multi-branch aggregation network with multiple stages to reconstruct sharp depth boundaries. Specifically, they design three parallel branches -- the reconstruction, the color, and the multi-scale branches -- in each stage. 
\cite{song2020channel} develop a coarse-to-fine framework consisting of several sub-modules that can gradually extract the high-frequency features from depth maps. Each sub-module employs a channel attention strategy to obtain informative features. Besides, the framework further improves the depth quality through Total Generalized Variation (TGV) term and input loss. 
In \cite{he2021towards}, a high-frequency guidance branch is developed to adaptively extract the HF information and reduce the LF components. The HF components are then fused with depth features by means of concatenation.
\cite{yuan2023recurrent} propose a {recurrent structure attention guidance framework (RSAG)} that utilizes both multi-scale and multi-stage strategies, shown in Fig.~\ref{fig:sr_RSAG}. In this framework, a deep contrastive network with multi-scale filters is designed to adaptively separate HF and LF features. 

Instead of deploying spatially-invariant kernels in vanilla convolutional neural networks (CNNs), some methods design learnable kernels to boost the GDSR performance.
\cite{kim2021deformable} propose a deformable network that provides a custom convolution kernel for each pixel by computing neighborhood weights.  The kernel weights can be obtained not only by calculating the pixels in regular positions (e.g., 8-neighborhood), but also by calculating its sub-pixels after interpolation.
\cite{wang2022learning} introduce a continuous depth representation whose core is the proposed geometric spatial aggregator (GSA). The GSA comprises two parts: (1) the geometric encoder using the scale-modulated distance field to build the correlation between pixels and (2) the learnable kernel learning typical texture pattern processing prior. Thanks to the continuous representation, the model can super-solve depth at an arbitrary scale.   
\cite{zhong2023deep} design a kernel generation network that can cope with inconsistent structures between RGB and depth.

To increase the interpretability, \cite{tang2021joint} regards GDSR as a neural implicit interpolation problem, which can map continuous coordinates of LR depth and HR color images into latent codes. Meanwhile, they learn the interpolation weights to establish a unified framework to yield the interpolation weights and values. 
{To promote the deployment of dToF cameras in mobile devices, \cite{li2022deltar} perform depth estimation from a lightweight ToF sensor and RGB image (DELTAR). In this work, PointNet~\citep{qi2017pointnet} and Efficient B5~\citep{tan2019efficientnet} are used to extract depth and RGB features, respectively, fused at the decoder stage through the proposed transformer-based fusion module.}

\subsection{Hybrid Architecture}
As neural networks are usually described as black-box models, some researchers combine optimization methods to obtain explainable frameworks.
\cite{riegler2016deep} present a novel network consisting of two sub-networks, i.e., a fully-convolutional network and a primal-dual network, with the first yielding HR depth maps and weights and the latter producing the final results with a non-local variational method.

As an optimization-inspired network, a weighted analysis sparse representation (WASR) model~\citep{gu2019learned} is designed for GDSR. It leverages a neural network to learn filter parameterizations and non-linear functions to build more flexible stage-wise operations.
Inspired by the multi-modal convolutional sparse coding model~\citep{song2019multimodal}, \cite{deng2020deep} develop a network that can adaptively separate the shared features between different modalities from the unique features existing in a single modality. 
\cite{marivani2020multimodal} propose a deep unfolding network to efficiently compute the convolutional sparse coding with the guide. 

\begin{figure}[t]
   \centering
    \begin{tabular}{c}
		\includegraphics[width=0.95\linewidth]{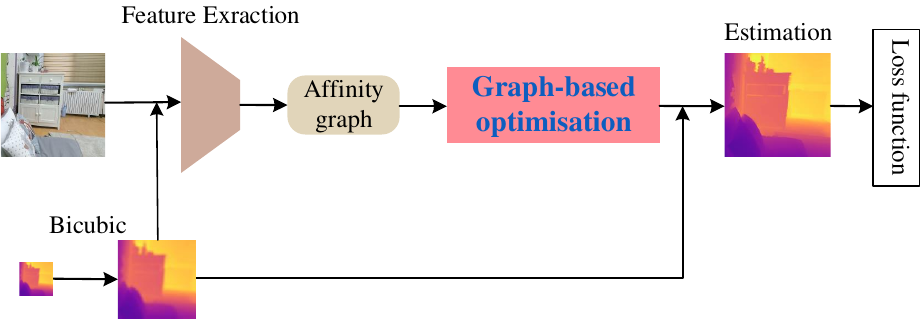} 
    \end{tabular}
    \caption{LGR architecture~\citep{de2022learning} {belonging to optimization-inspired architecture.} The optimization layer is plugged into the end of the network to predict the high-resolution depth.}
    \label{fig:sr_LGR}
\end{figure}

\cite{zhao2022discrete} introduce discrete cosine transform (DCT) into the proposed neural network to make the model explainable. This model utilizes DCT to tackle the optimization issue for GDSR by reconstructing the multi-channel HR depth features. Another advantage of using DCT is that it can make network design easier because it does not need to learn the mapping function between LR/HR image pairs, as it explicitly models it.  
\cite{de2022learning} propose a novel architecture {learning graph regularization (LGR)} with a neural network to perform GDSR, as shown in Fig.~\ref{fig:sr_LGR}. First, a network extracts features from the source and the guide. A graph constructed using such features is then sent to a differentiable optimization layer to regularize it.
Combining the advantages of model-based and learning-based methods, \cite{zhou2023memory} propose a memory-augmented deep unfolding network in which local implicit prior and global implicit prior are introduced based on a maximal posterior view. To further prevent information loss, long short-term unit (LSTM) is employed in the persistent memory mechanism.
\cite{metzger2022guided} propose  integrating guided anisotropic diffusion into a CNN to enhance depth discontinuities in the depth images. It has a fixed inference time and memory footprint, given a scale factor.

\subsection{Auxiliary-based Methods}       \label{aux_gdsr}
\begin{figure}[t]
   \centering
    \begin{tabular}{c}
		\includegraphics[width=0.95\linewidth]{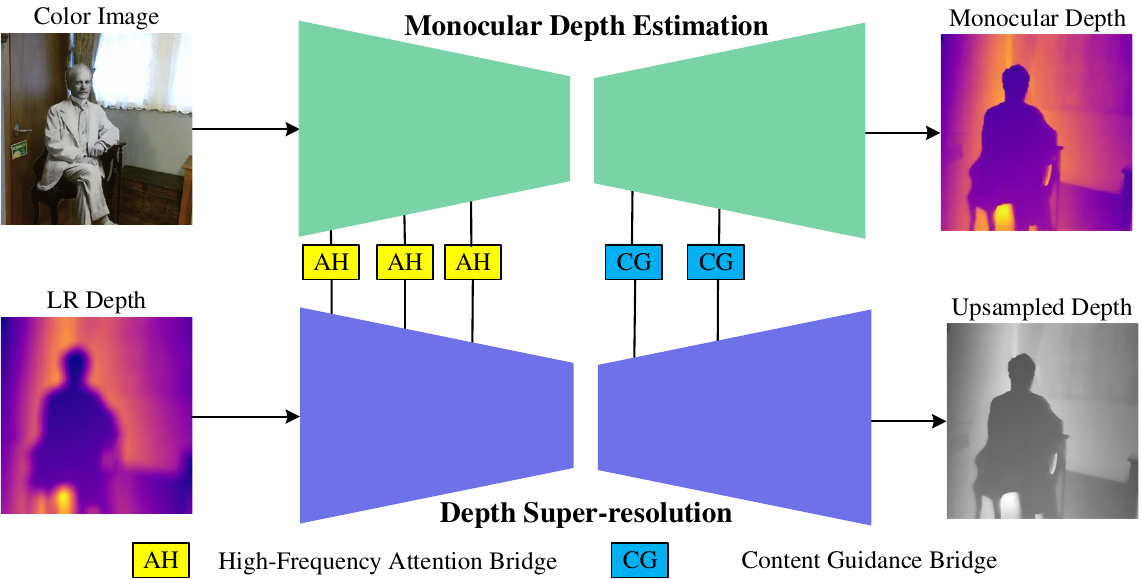} 
    \end{tabular}
    \caption{BridgeNet architecture~\citep{tang2021bridgenet} {belonging to auxiliary-based methods.} The BridgeNet consists of a monocular depth estimation subnetwork and a guided depth upsampling subnetwork.}
    \label{fig:sr_bridgenet}
\end{figure}

Auxiliary learning, also known as multi-task learning, aims to find or design auxiliary tasks that can improve the performance of one or several primary tasks. Driven by its recent progress, several works attempt to exploit auxiliary task learning to upsample depth maps. 
\cite{sun2021learning} propose a knowledge distillation approach where depth estimation is employed as an auxiliary task during training to improve the results. 
As a representative method shown in Fig.~\ref{fig:sr_bridgenet}, BridgeNet~\citep{tang2021bridgenet} explores a paradigm where the association between monocular depth estimate (MDE) and depth super-resolution (DSR) is exploited deeply. To make the two tasks work together, they design two auto-encoders, namely DSRNet and MDENet, with information interaction at the encoder stage. In MDENet, a high-frequency attention bridge is proposed to capture HF information which can guide depth upsampling in the other task. In contrast, the content guidance is provided by DSRNet to guide monocular depth estimates. Considering the distinct learning curves of the two sub-tasks, they optimize the two sub-networks separately.

In addition to auxiliary learning, some researchers consider using auxiliary guidance from color images to improve depth quality. 
Based on the importance of edge information in depth upsampling, \cite{wang2020depth} propose an edge-guided depth upsampling framework that leverages edge maps as the guide to recover HR depth maps. For training, the ground truth edge map is computed by the Canny operator on the corresponding HR depth map.  

Differently from most existing GDSR methods, \cite{voynov2019perceptual} attempt to gauge the upsampled depth quality through renderings of surface normal maps. More specifically, a visual appearance-based loss is proposed, which can assist a baseline network in yielding more visually pleasing results.

\subsection{\revise{Fusion-focused Methods}}
How to effectively fuse information from different modalities is critical to obtain high-quality depth maps. Intuitively, we can perform multi-modal fusion with a simple operation, such as concatenation~\citep{wang2020depth,tang2021joint,de2022learning}. However, these fusion schemes can not selectively transfer HF features from the guide to the target, and may even result in texture-copying artifacts. Therefore, researchers design different fusion schemes to alleviate the issue, mainly grouped into three categories: early, late, and multi-level fusion. 
\cite{li2020depth} present a correlation-controlled color guidance block (block) to fuse the multi-modal information.
\cite{zhong2021high} design a multi-modal attention-based fusion strategy, including a feature enhancement block and a feature recalibration block, with the former paying more attention to meaningful features extracted from depth maps and color images and the latter aiming at rescaling multi-modal features. This strategy can effectively avoid the texture-copying effect in the final prediction.
\cite{zhao2022discrete} deploy an enhanced spatial attention block to transfer meaningful structural information from RGB to depth. 
\cite{yuan2023recurrent} develop a recurrent structure attention block, where the latest depth estimate and its corresponding HR color image are taken as the input, to obtain useful HF features of the color image.
\cite{wang2023joint} propose a multi-perspective cross-guided fusion filter block, to improve depth features gradually by fusing the structure details in RGB images. In this block, spatial representations from various views are learned to capture depth salient structures further. Moreover, a color-depth cross-attention module is employed to achieve edge preservation. 
\cite{zhong2023deep} propose a multi-scale guided filtering module to refine the depth map in a coarse-to-fine manner.
{To obtain stable depth predictions, \cite{sun2023consistent} conduct cross-modal fusion in a multi-frame manner, i.e., dToF depth video super-resolution (DVSR) and histogram video super-resolution (HVSR), to upsample LR depth maps.}
\revise{\cite{qiao2023depth} propose an Adaptive Feature Fusion Module (AFFM), which enables the recovery of fine details from the HR guide and the LR depth maps.
\cite{zhao2023spherical} propose a Spherical Space feature Decomposition network (SSDNet), which projects encoded features onto the spherical space. This strategy allows for separating and aligning domain-shared and domain-private features, respectively.}

\subsection{Unsupervised Methods}
\begin{figure}[t]
   \centering
    \begin{tabular}{c}
		\includegraphics[width=0.95\linewidth]{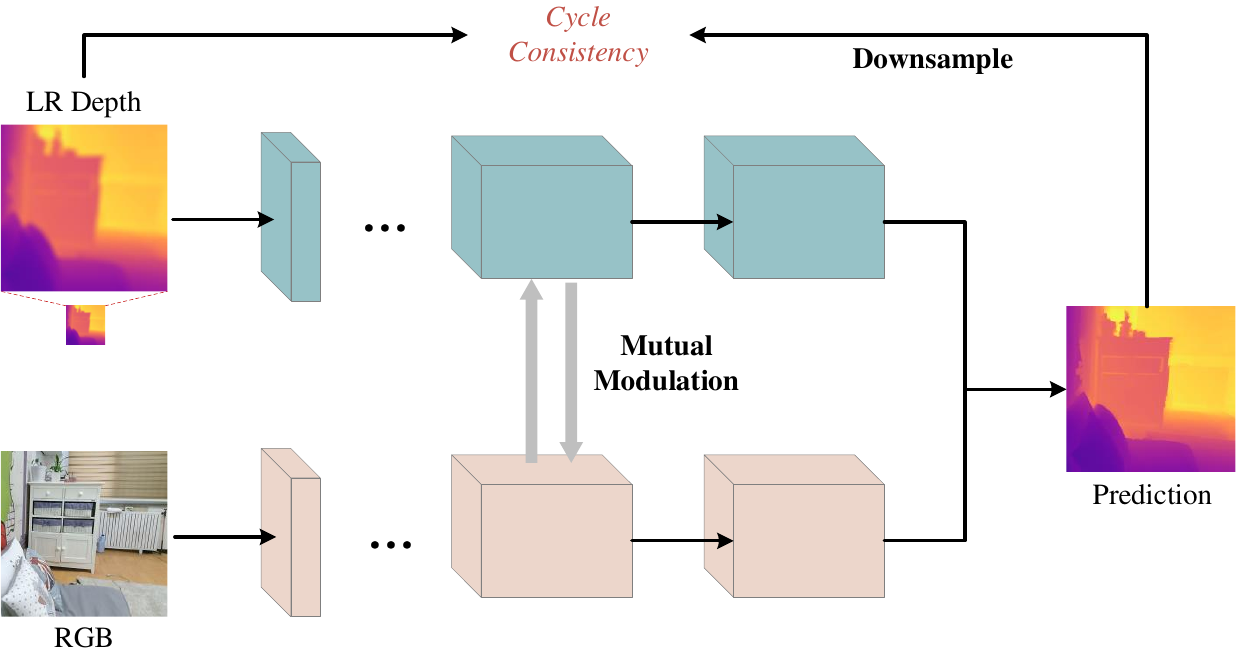} 
    \end{tabular}
    \caption{MMSR framework \citep{dong2022learning} {belonging to unsupervised methods.} The MMSR adopts the cycle consistency loss to train the network in a self-supervised manner.}
    \label{fig:sr_MMSR}
\end{figure}

Since collecting training data paired with annotations is labor-intensive and time-consuming, researchers seek to solve the issue by leveraging self-supervised learning, allowing training models without labels.
In \cite{lutio2019guided}, GDSR is regarded as a pixel-wise mapping from the source to the target, implemented as a multi-layer perceptron. This formulation can be trained in a fully unsupervised manner with the constraint of having the LR source.
\cite{shacht2021single} present a single-pair method that can upsample depth maps even when the input pairs are misaligned. During training, patches cropped from the input pairs are utilized as pseudo-label data to enable weakly supervised learning. 
To improve generalization, \cite{dong2022learning} propose a mutual modulation super-resolution model {(MMSR)}, where a cross-modality modulation strategy using adaptive filters transfers meaningful features from one modality to the other, as shown in Fig.~\ref{fig:sr_MMSR}. In this mutual modulation, the spatial relationship between the corresponding pixels of the two modalities is fully exploited. Moreover, a cycle consistency loss is adopted to enforce the target faithful to the source.
Differently from previous unsupervised methods for GDSR, \cite{shin2023task} introduce a pretext task into this field. Through the proposed scene structure guidance network, explicit structure features of color images can be obtained, which, together with the corresponding LR depth maps, serve as the input of a baseline network to produce high-resolution depth maps.
\cite{wang2023self} construct an adversarial network where the dependency between RGB images and depth maps is fully exploited to enhance the depth. Further, the optimal transport theory is introduced into the framework to boost depth enhancement performance.
\revise{\cite{qiao2023self} exploit a contrast learning scheme to extract unique features from the guidance, which can boost the GDSR performance based on a baseline network.}

To conclude this section, Tab. \ref{gdsr_list} collects any of the methods discussed so far, divided into five categories.  For each method, the venue and year are reported, together with a short description of the key idea behind it.

\begin{table*}[]
\centering
\scalebox{0.5}{
\setlength{\tabcolsep}{3pt}{
\begin{tabular}{lllcccc}
\hline
Paradigm    & Method    & Reference & \multicolumn{4}{c}{Key Idea} \\ \hline

\multirow{30}{*}{\rotatebox[origin=b]{90}{\textbf{Dual-branch Framework}}}

&  DJFR~\citep{li2019joint}      &  TPAMI-2019  &\multicolumn{4}{l}{\begin{tabular}[l]{@{}l@{}}Introduce two sub-networks for extracting features from the target and guidance images. Following the \\concatenation operation, the fused information is fed into another sub-network for the predicted depth.\end{tabular}}   \\ 

&  DMSG~\citep{hui2016depth}      &   ECCV-2016      & \multicolumn{4}{l}{\begin{tabular}[l]{@{}l@{}} First work to super-solve depth maps using multi-scale guidance, which progressively enhances depth details from\\ high- to low-level.\end{tabular}} \\ 

&  DepthSR~\citep{guo2018hierarchical}   &  TIP-2019       &  \multicolumn{4}{l}{\begin{tabular}[l]{@{}l@{}}Exploit a residual U-Net structure, where hierarchical guidance is used at each scale, to recover HR depth maps.\end{tabular}}   \\

&  MGD~\citep{zuo2019depth}     &  TCSVT-2019     &  \multicolumn{4}{l}{\begin{tabular}[l]{@{}l@{}}Propose a multi-scale architecture based on dense layers that aim to revisit the features from all higher levels at a \\given scale.\end{tabular}}  \\ 

&  MFR-SR~\citep{zuo2019multi}    &  TCSVT-2019      &  \multicolumn{4}{l}{\begin{tabular}[l]{@{}l@{}}Combine global and local residual learning to upsamples depth maps from
coarse to fine via multi-scale frequency\\ synthesis.\end{tabular}}  \\ 

&  MIG-net~\citep{zuo2021mig}     &  TMM-2021      &\multicolumn{4}{l}{\begin{tabular}[l]{@{}l@{}} Introduce a two-stream multi-scale network to extract features in the image and gradient domains, where the depth \\features and gradient features are alternatively complemented with each other.\end{tabular}}    \\ 

&  SFG~\citep{yuan2023structure}     &  AAAI-2023        &\multicolumn{4}{l}{\begin{tabular}[l]{@{}l@{}} Propose a structure flow-guided framework for GDSR, whose key point is to learn a structure flow map to guide\\structure representation transferring.\end{tabular}}   \\ 

&  CCFN~\citep{wen2018deep}    &  TIP-2019        &\multicolumn{4}{l}{\begin{tabular}[l]{@{}l@{}}Propose a coarse-to-fine neural network with two stages. In the first stage, the model learns large filter kernels to\\ obtain coarse results, which are further refined by smaller filter kernels in the second stage.\end{tabular}} \\  

&  CGN~\citep{zuo2020frequency}    & TMM-2020      &\multicolumn{4}{l}{\begin{tabular}[l]{@{}l@{}}With the proposed depth-guided affine transformation, the intensity-guided depth feature filtering and refinement \\are carried out iteratively in multiple stages.\end{tabular}} \\    

&  \multirow{2}{*}{PMBANet~\citep{ye2020pmbanet}}    &   \multirow{2}{*}{TIP-2020}    & \multicolumn{4}{l}{\multirow{2}{*}{\begin{tabular}[l]{@{}l@{}}Introduce a multi-branch aggregation network with multiple stages to reconstruct sharp depth boundaries.\end{tabular}}}  \\ 
&   &   & \\

&\multirow{2}{*}{IRLF~\citep{song2020channel}}      
&\multirow{2}{*}{CVPR-2020}         
&\multicolumn{4}{l}{\multirow{2}{*}{\begin{tabular}[l]{@{}l@{}}Develop a coarse-to-fine framework consisting of several sub-modules with channel attention strategies.
\end{tabular}}}    \\
&   &   &  \\

&\multirow{2}{*}{DKN~\citep{kim2021deformable}}       
&\multirow{2}{*}{IJCV-2021}         
&\multicolumn{4}{l}{\multirow{2}{*}{\begin{tabular}[l]{@{}l@{}}Design a deformable kernel network that obtains a kernel for each pixel based on the neighborhood weights.\end{tabular}}}  \\
&   &   &  \\ 

&\multirow{2}{*}{FDSR~\citep{he2021towards}}      
&\multirow{2}{*}{CVPR-2021}          
&\multicolumn{4}{l}{\multirow{2}{*}{\begin{tabular}[l]{@{}l@{}}
Explore high-frequency (HF) information via an HF guidance branch, where the HF information is fused with depth\\ features to improve the performance.\end{tabular}}}   \\
&   &   &   \\

&\multirow{2}{*}{GeoDSR~\citep{wang2022learning}}    
&\multirow{2}{*}{AAAI-2023}         
&\multicolumn{4}{l}{\multirow{2}{*}{\begin{tabular}[l]{@{}l@{}}Introduce a continuous depth representation to effectively implement scale-continuous and spatial-continuous\\ upsampling in guided depth super-resolution.\end{tabular}}}    \\
&   &   &  \\

&\multirow{2}{*}{DAGF~\citep{zhong2023deep}}      
&\multirow{2}{*}{TNNLS-2023}         
&\multicolumn{4}{l}{\multirow{2}{*}{\begin{tabular}[l]{@{}l@{}}Design a kernel generation network that can cope with inconsistent structures between RGB and depth, where a \\multi-scale guided filtering module is proposed to refine the depth map in a coarse-to-fine manner.\end{tabular}}}   \\
&   &   &  \\ 

&\multirow{2}{*}{DCSR~\citep{wang2023depth}}      
&\multirow{2}{*}{Display-2023}          
&\multicolumn{4}{l}{\multirow{2}{*}{\begin{tabular}[l]{@{}l@{}} Proposes a depth map continuous SR framework that can achieve resolution adaptation at arbitrary SR ratios.\end{tabular}}}   \\   
&   &   &     \\ 

&\multirow{2}{*}{PAC~\citep{su2019pixel}}      
&\multirow{2}{*}{CVPR-2019}          
&\multicolumn{4}{l}{\multirow{2}{*}{\begin{tabular}[l]{@{}l@{}} Present a pixel-adaptive convolution operation for deep joint image upsampling.\end{tabular}}}   \\   
&   &   &     \\ 

&\multirow{2}{*}{DELTAR~\citep{li2022deltar}}      
&\multirow{2}{*}{ECCV-2022}          
&\multicolumn{4}{l}{\multirow{2}{*}{\begin{tabular}[l]{@{}l@{}} Propose a transformer-based architecture with two branches, which can extract and fuse depth and RGB features \\ efficiently. Besides, a cross-modal calibration is performed to align RGB and depth.  \end{tabular}}}   \\   
&   &   &     \\  \hline

\multirow{16}{*}{\rotatebox[origin=b]{90}{\textbf{Hybrid Model}}}

& \multirow{2}{*}{DPDN~\citep{riegler2016deep}}       &   \multirow{2}{*}{BMVC-2016}      & \multicolumn{4}{l}{\multirow{2}{*}{\begin{tabular}[l]{@{}l@{}}Use two sub-networks. one based on a fully-convolutional structure produces HR depth maps and weights.\\ Another uses a non-local variational method to get final results.\end{tabular}}}   \\ 
&   &   & \\  

&\multirow{2}{*}{DG-CMM~\citep{gu2019learned}} &\multirow{2}{*}{TPAMI-2019}           
&\multicolumn{4}{l}{\multirow{2}{*}{\begin{tabular}[l]{@{}l@{}}Propose a weighted analysis sparse
representation (WASR) model and a neural network to learn filter \\parameterizations and non-linear functions to build flexible stage-wise operations.\end{tabular}}}   \\
&   &   & \\ 
                        
&\multirow{2}{*}{CUNet~\citep{deng2020deep}}   
&\multirow{2}{*}{TPAMI-2020}         
&\multicolumn{4}{l}{\multirow{2}{*}{\begin{tabular}[l]{@{}l@{}}Develop a multi-modal convolutional
sparse coding (MCSC) models to solve the general multi-modal image\\ restoration and multi-modal image fusion problems.\end{tabular}}}  \\ 
&   &   &   \\ 

&\multirow{2}{*}{LMCSC~\citep{marivani2020multimodal}}     
&\multirow{2}{*}{TIP-2020}         
&\multicolumn{4}{l}{\multirow{2}{*}{\begin{tabular}[l]{@{}l@{}}Propose a deep unfolding network using guided
image SR that exploits information from two modalities.\end{tabular}}}   \\
&   &   &  \\ 

&\multirow{2}{*}{LGR~\citep{de2022learning}}       
&\multirow{2}{*}{CVPR-2022}            
&\multicolumn{4}{l}{\multirow{2}{*}{\begin{tabular}[l]{@{}l@{}}Propose a novel architecture combining a neural network and graph-based optimization to perform GDSR.\end{tabular}}}   \\
&   &   &   \\ 

&\multirow{2}{*}{DCTNet~\citep{zhao2022discrete}}    
&\multirow{2}{*}{CVPR-2022}   
&\multicolumn{4}{l}{\multirow{2}{*}{\begin{tabular}[l]{@{}l@{}}Utilizes discrete cosine transform (DCT) to tackle the optimization issue for GDSR by reconstructing the\\ multi-channel HR depth features.\end{tabular}}}   \\
&   &   &   \\  

&\multirow{2}{*}{DaDa~\citep{metzger2022guided}}      
&\multirow{2}{*}{CVPR-2023}          
&\multicolumn{4}{l}{\multirow{2}{*}{\begin{tabular}[l]{@{}l@{}} Suggest incorporating guided anisotropic diffusion into a CNN to improve depth discontinuities in depth images.\\ Given a scale factor, the inference time and memory demands of the method are fixed.\end{tabular}}}   \\
&   &   &  \\ 

&\multirow{2}{*}{MADUNet~\citep{zhou2023memory}}   
&\multirow{2}{*}{IJCV-2023}         
&\multicolumn{4}{l}{\multirow{2}{*}{\begin{tabular}[l]{@{}l@{}} Propose a memory-augmented deep unfolding network where the introduction of local and global implicit priors is\\ based on a maximal a posterior view.\end{tabular}}}  \\
&   &   &  \\ \hline

\multirow{12}{*}{\rotatebox[origin=b]{90}{\textbf{Auxiliary-based}}}

&\multirow{2}{*}{CTKT~\citep{sun2021learning}}           
&\multirow{2}{*}{CVPR-2021}          
&\multicolumn{4}{l}{\multirow{2}{*}{\begin{tabular}[l]{@{}l@{}} Propose a knowledge distillation method that uses the depth estimate auxiliary task during training for better results.\end{tabular}}}  \\ 
&   &   &   \\ 

&\multirow{2}{*}{BridgeNet~\citep{tang2021bridgenet}} 
&\multirow{2}{*}{ACM MM-2021}          
&\multicolumn{4}{l}{\multirow{2}{*}{\begin{tabular}[l]{@{}l@{}} Explore a paradigm where the association between monocular depth estimate (MDE) and depth super-resolution \\ are used to advance the performance of depth map super-resolution.\end{tabular}}}   \\
&   &   &  \\ 

& \multirow{2}{*}{DSR-N~\citep{wang2020depth}}     &\multirow{2}{*}{PR-2020}  
& \multicolumn{4}{l}{\multirow{2}{*}{\begin{tabular}[l]{@{}l@{}} Propose a depth upsampling framework that leverages edge information as the guide to recover HR depth maps.\end{tabular}}}  \\
&   &   &   \\

&\multirow{2}{*}{PDSR~\citep{voynov2019perceptual}} &\multirow{2}{*}{ICCV-2019}         
&\multicolumn{4}{l}{\multirow{2}{*}{\begin{tabular}[l]{@{}l@{}} Attempt to measure the upsampled depth quality using renderings of surface normal maps.\end{tabular}}}  \\ 
&   &   &   \\

&\multirow{2}{*}{PDR-Net~\citep{liu2022pdr}}   
&\multirow{2}{*}{Neurocomputing-2022}
&\multicolumn{4}{l}{\multirow{2}{*}{\begin{tabular}[l]{@{}l@{}} Proposes a progressive depth reconstruction network to further enhance the performance of DMSR.\end{tabular}}}   \\
&   &   &   \\ 

&\multirow{2}{*}{SVLRM~\citep{dong2021learning}}   
&\multirow{2}{*}{TPAMI-2021}
&\multicolumn{4}{l}{\multirow{2}{*}{\begin{tabular}[l]{@{}l@{}} Propose a new joint filtering method based on a spatially variant linear
representation model where the target image is \\linearly represented by the guidance image.\end{tabular}}}   \\
&   &   &   \\ \hline

\multirow{14}{*}{\rotatebox[origin=b]{90}{\textbf{Fusion-focused}}}
                                                                
&\multirow{2}{*}{JIIF~\citep{tang2021joint}}    
&\multirow{2}{*}{MM-2021}          
&\multicolumn{4}{l}{\multirow{2}{*}{\begin{tabular}[l]{@{}l@{}}Considers GDSR to be a neural implicit interpolation problem that can convert latent codes from continuous\\ coordinates in LR depth and HR color images.\end{tabular}}}   \\
&   &   &   \\ 

&  MSSN~\citep{li2020depth}    &  PR-2020       &\multicolumn{4}{l}{\begin{tabular}[l]{@{}l@{}} Present a multi-scale strategy with the proposed symmetric unit as the basic component of the network, which can\\ effectively deal with textureless and edge features in depth images.\end{tabular}}    \\ 

&  AHMF~\citep{zhong2021high}     &  TIP-2021         &\multicolumn{4}{l}{\begin{tabular}[l]{@{}l@{}} Propose a hierarchical network based on a bi-directional hierarchical feature collaboration module, where low- and\\ high-level spatial information can work together to improve each other.\end{tabular}}   \\ 

&\multirow{2}{*}{RSAG~\citep{yuan2023recurrent}}      
&\multirow{2}{*}{AAAI-2023}          
&\multicolumn{4}{l}{\multirow{2}{*}{\begin{tabular}[l]{@{}l@{}}Propose a recurrent structure attention guidance framework that employs both multi-scale and multi-stage \\strategies. A deep contrastive network with multi-scale filters is used to separate HF and LF features.\end{tabular}}}    \\
&   &   &  \\ 

&  JGF~\citep{wang2023joint}      &  PR-2023         &\multicolumn{4}{l}{\begin{tabular}[l]{@{}l@{}} Present a novel method that uses pyramid structure to capture multi-scale features, allowing HR depth maps to be\\ recovered progressively.\end{tabular}}   \\ 

& DVSR~\citep{sun2023consistent}    
& CVPR-2023       
& \multicolumn{4}{l}{\begin{tabular}[l]{@{}l@{}} Propose a multi-frame scheme to upsample LR dToF videos with corresponding HR RGB sequences. \end{tabular}}    \\  

&\multirow{2}{*}{\revise{SSDNet~\citep{zhao2023spherical}}}      
&\multirow{2}{*}{\revise{ICCV-2023}}         
&\multicolumn{4}{l}{\multirow{2}{*}{\begin{tabular}[l]{@{}l@{}} \revise{Propose a spherical contrast refinement module to enhance depth details from RGB features.} \end{tabular}}}   \\
&   &   &  \\  

&\revise{DSR-EI~\citep{qiao2023depth}}      
&\revise{CVIU-2023}         
&\multicolumn{4}{l}{\begin{tabular}[l]{@{}l@{}} \revise{Propose AFFM to adaptively fuse discriminative cross-modality features.} \end{tabular}}   \\
&   &   &  \\  \hline

\multirow{11}{*}{\rotatebox[origin=b]{90}{\textbf{Unsupervised}}} 
&  P2P~\citep{lutio2019guided}     
&  ICCV-2019         
&  \multicolumn{4}{l}{\begin{tabular}[l]{@{}l@{}} GDSR is regarded as a pixel-wise mapping from source to target and implemented as a multi-layer perceptron.\end{tabular}}   \\
                        
&  CMSR~\citep{shacht2021single}     
&  CVPR-2021           
&  \multicolumn{4}{l}{\begin{tabular}[l]{@{}l@{}} Present a single-pair method that can upsample depth maps even when the input pairs are misaligned, where\\ patches cropped from the input pairs are utilized as pseudo-label data to enable weakly supervised learning.\end{tabular}}   \\ 

&  MMSR~\citep{dong2022learning}      
&  ECCV-2022        
&  \multicolumn{4}{l}{\begin{tabular}[l]{@{}l@{}} A cross-modality modulation strategy using adaptive filters is developed to transfer significant features from one\\ modality to the other in a proposed mutual modulation super-resolution model.\end{tabular}}   \\ 
                              
&  DEDE~\citep{wang2023self}   
&  TIP-2023            
&  \multicolumn{4}{l}{\begin{tabular}[l]{@{}l@{}} Design an adversarial network where the relationship between depth maps and RGB images are fully exploited\\ to improve the depth.\end{tabular}}    \\ 

&  SSGnet~\citep{shin2023task}   
&  AAAI-2023        
&  \multicolumn{4}{l}{\begin{tabular}[l]{@{}l@{}} Explicit structure features of color images can be extracted using the suggested scene structure guidance network, \\and these features—along with the corresponding LR depth maps—serve as the input of a base network.\end{tabular}}    \\ 

&  \revise{CMPNet~\citep{qiao2023self}}
&  \revise{NN-2023}        
&  \multicolumn{4}{l}{\begin{tabular}[l]{@{}l@{}} \revise{Develop an autoencoder-based framework
with contrastive and reconstruction losses} \\ \revise{to reduce information redundancy.}\end{tabular}}    \\  \hline
\end{tabular}}
}
 \caption{List of Guided Depth Super-Resolution methods, divided according to the five outlined categories.}
 \label{gdsr_list}
\end{table*}

\section{Guided Depth Completion} \label{dcp}
Pivotal works on depth completion~\citep{uhrig2017sparsity,chodosh2019deep} only take depth maps as input. However, due to the sparsity of depth maps obtained from long-range LiDAR measurements, a substantial amount of details, such as object boundaries, are lost. Thus, restoring these details is difficult on densified depth maps, even using deep learning. Therefore, researchers started exploiting additional modalities capturing the same scene to guide the completion of sparse depth maps. In practice, color images can provide rich texture, scene structure, and object details, thus allowing for the extraction of sufficient cues to enhance depth completion. Besides, color images can also offer various auxiliary cues, such as semantics, monocular depth, and surface normal. {Similar to GDSR}, we divide GDC into two categories: supervised and unsupervised methods. Supervised methods can be further classified into dual-branch architectures, spatial propagation networks, auxiliary-based methods, and uncertainty guidance. These six categories are listed in the bottom branch of Fig. \ref{method_overview}.


Most existing approaches employ supervised learning for this task, which typically constructs image pairs from dense maps and their sparse version for training. With dense depth information being used as annotation, deep neural networks are trained to solve a regression problem, as done for GDSR.

\subsection{Dual-branch Architecture}
The dual-branch architecture is common in GDC. It utilizes two encoders to extract features from RGB and sparse depth maps, and then the cross-modal information is sent to a single decoder after aggregation. In this latter, a designed information fusion strategy can also be incorporated. Thus, two folds must be focused on: effective representation learning and information fusion. 

\cite{jaritz2018sparse} employ a dual-branch architecture based on a modified version of {the neural architecture search network (NASNet)}~\citep{zoph2018learning}. First, the features from RGB images and depth maps are extracted using two NASNet encoders. Then, the intermediate features are fused in a late fusion scheme through channel-wise concatenation and fed into a decoder, yielding the final dense estimate. In this framework, validity masks are demonstrated to fail further to improve performance for GDC in large neural networks. To address the issue of model performance degradation caused by blurry guidance in the color images and unclear structures in the depth maps, \cite{yan2022rignet} exploit a repetitive design based on the hourglass network~\citep{tang2020learning}.  
To achieve multi-scale training, \cite{li2020multi} introduce the cascaded architecture into the hourglass network, extended by \cite{Fan_2022_BMVC} with denser interaction between the two modalities.    

\begin{figure}[t]
   \centering
    \begin{tabular}{c}
		\includegraphics[width=0.95\linewidth]{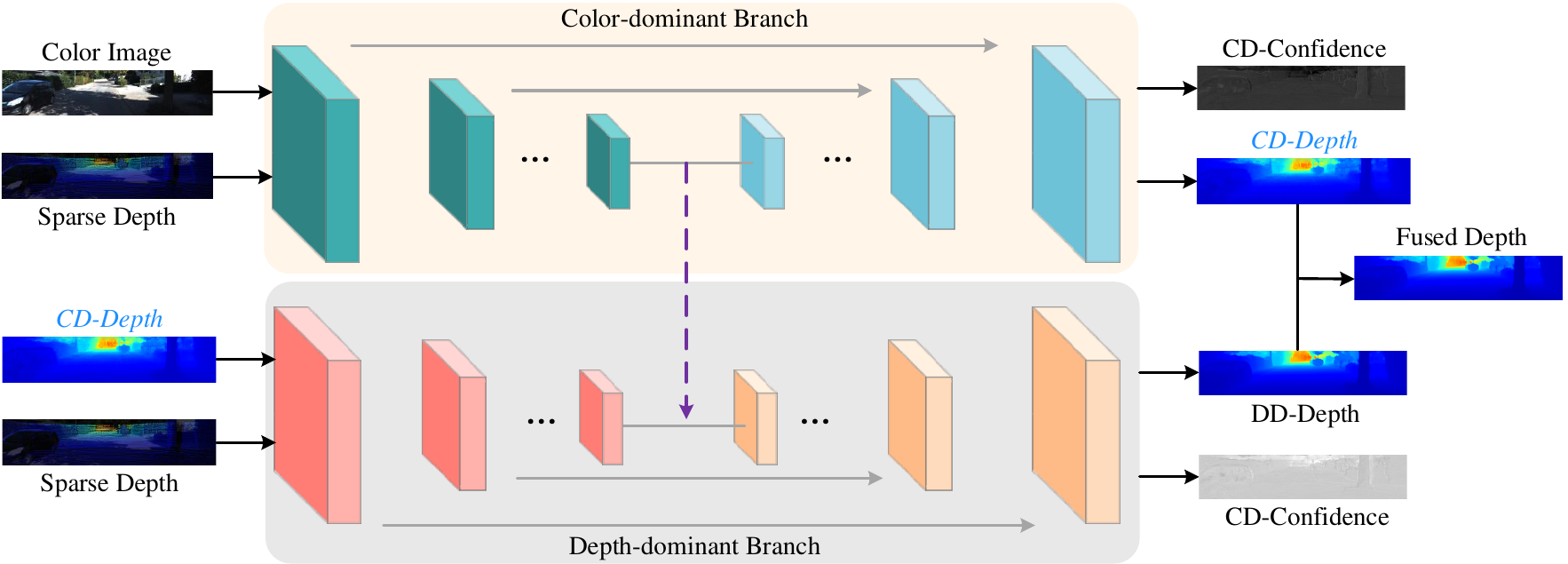} 
    \end{tabular}
    \caption{PENet~\citep{hu2021penet} framework {belonging to dual-branch architectures.} It consists of two branches that can respectively produce depth estimates from color-dominant and depth-dominant information.}
    \label{fig:sr_penet}
\end{figure}

In the two-branch framework proposed by~\cite{van2019sparse}, global and local information are extracted with two encoders based on ERFNet~\citep{romera2017erfnet}, respectively. Notably, the local branch uses two U-nets to reconstruct sharp edges and structural details. Finally, the predictions are weighted by the outputs of each branch. 
\cite{hu2021penet} propose {a network that can perform precise and efficient depth completion (PENet)}, as depicted in Fig.~\ref{fig:sr_penet}. It gradually enhances the accuracy of depth prediction in a coarse-to-fine manner. 
The first branch, named the color-dominant branch, takes a color image and a sparse depth map as input to produce a dense depth map that inherits sharp edges and structural details in the color image. The second branch, i.e., the depth dominant branch, employs a similar network, whose inputs are the previous dense and sparse depth map, to output another dense map. Finally, the two depth maps are fused using the strategy proposed by FusionNet~\citep{van2019sparse}. Notably, they concatenate the 3D position maps to the convolutional layer, forming a geometric convolutional layer, to lift the performance further. 
For indoor depth completion, \cite{jiang2022low} use a dual-branch structure to achieve an efficient solution with low power consumption.

Using a dense pseudo-depth map as a coarse reference for the final prediction, \cite{gu2021denselidar} can output stable and accurate dense depth maps.
Different from the vanilla point-estimate methods with a dual-branch architecture, \cite{yang2019dense} develop a conditional prior network (CPN) to calculate a posterior probability over the depth of each pixel, combined with a likelihood term using sparse measurements.
The standard convolution operation applies a kernel to the pixel grid in an image, which, however, cannot handle sparse depth maps with uneven distribution of depth values. To address this issue, \cite{zhao2021adaptive} propose to leverage graph propagation to extract spatial contexts. Considering the variability of the graph structure, it applies the co-attention module~\citep{lu2016hierarchical} on the guided graph propagation for efficient multi-modal representation extraction. Combining traditional methods with a dual-branch deep network, \cite{xu2023real} achieve promising results.
In the most recent works, \cite{chen2023agg} employ two parallel Unet-shape networks to extract RGB and depth features at different scales. Then, feature fusion is performed in the decoding stage through the proposed attention-guided skip connection (AG-SC) module, yielding the desired results.
In contrast, \cite{yu2023aggregating} propose a network named PointDC that leverages 2D and 3D information to conduct robust depth completion.

\textbf{Guided image filtering.} In GDC, deep-guided image filtering can be viewed as a type of dual-branch architecture that utilizes features from RGB images to predict a kernel. This kernel is often learned by a deep network and applied to depth for feature fusion or dense depth prediction. 

Inspired by classical guided image filtering~\citep{tomasi1998bilateral,he2012guided}, \cite{tang2020learning} propose a guided network consisting of a GuideNet and a DepthNet to recover dense depth maps. Inside the GuideNet, content-dependent and spatial-variant convolution kernels are predicted to capture geometric structural features in color images consistent with real-world scenes. Furthermore, a convolution factorization is introduced into the method to reduce computational overhead. 

\cite{liu2021learning} present a two-stage model where depth interpolation and refinement are performed sequentially. Specifically, they first interpolate the sparse depth map via the proposed differentiable kernel regression layer, proving more effective than hand-designed filters. Then, a residual network based on U-net is adopted to refine the coarse depth map.

\subsection{Affinity-based Methods}
In the form of a generic matrix, the affinity matrix expresses the pairwise proximity information between a set of data points. Since the spatial propagation network (SPN) for learning the affinity matrix was first employed for depth completion~\citep{liu2017learning}, many significant efforts have been made to improve it, yielding impressive results. In this pioneering work, the affinity matrix, which can establish associations between any two pixels over a whole image, is composed of the spatial transformation matrix. Given a hidden representation $H^{s}$ at iteration $t$, the spatial propagation with affinity matrix $w$ can be expressed as:
\begin{equation}
    \mathbf{H}_{p,q}^{t+1}=(I-d_{t})\mathbf{H}_{p,q}^{t}+\sum_{p,q \in N_{p,q}} w_{p,q}^{i,j} \mathbf{H}_{i,j}^{t}
\end{equation}
where $(p,q)$ and $(i,j)$ are the positions of reference and neighbor pixels, respectively, $N_{p,q}$ denote the neighborhood of $(p,q)$, and $d_{t}=1-\sum_{i,j\in N_{p,q}}w_{p,q}^{i,j}$.

\begin{figure}[t]
   \centering
    \begin{tabular}{c}
		\includegraphics[width=0.95\linewidth]{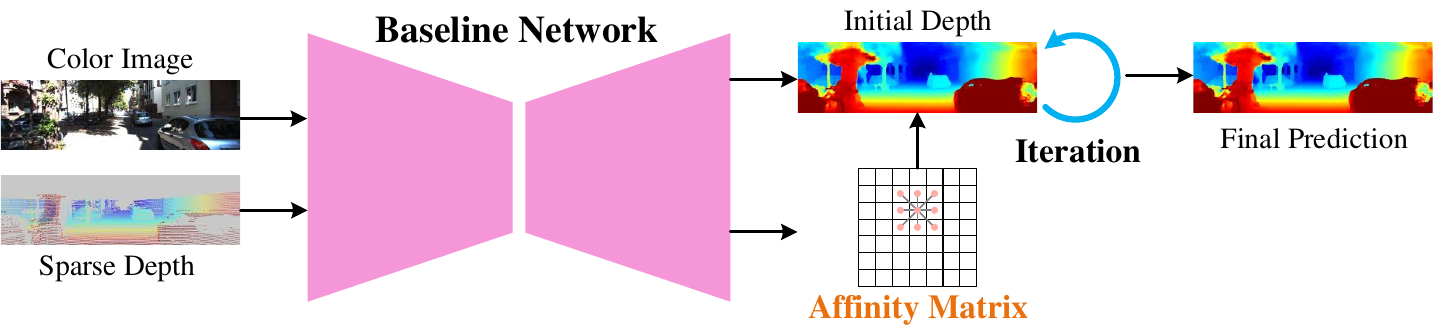} 
    \end{tabular}
    \caption{CSPN framework~\citep{cheng2018depth} {belonging to affinity-based methods}. At each step, CSPN propagates a local region in all directions at the same time.}
    \label{fig:sr_cspn}
\end{figure}

Since spatial propagation is only performed along a certain direction at a time, the time cost of this method, which is run in sequence, is high. 
In contrast, \cite{cheng2018depth} formulate the task as anisotropic diffusion filtering~\citep{weickert1998anisotropic,liu2016learning}, and implement it through a convolutional network named CSPN, depicted in Fig. \ref{fig:sr_cspn}. In detail, they first employ an Unet-based network that inputs an RGB image and a sparse depth map, and outputs a blurred depth map and an affinity matrix. Then, the propagation is carried out with recurrent convolutional operation. However, both approaches fix the receptive field in the propagation, which may yield suboptimal results for the task. To solve the problem, some methods utilizing flexible propagation schemes, such as {CSPN++~\citep{cheng2020cspn++}, non-local SPN (NLSPN)~\citep{park2020non}, deformable SPN (DSPN)~\citep{xu2020deformable} and dynamic SPN (DySPN)~\citep{lin2022dynamic}}, are proposed. 

To further improve the model performance for depth completion, \cite{cheng2020cspn++} propose CSPN++, which includes two variants, i.e., context-aware CSPN (CA-CSPN) and resource-aware CSPN (RA-CSPN). By converting two constants in CSPN -- the number of iterations and kernel size -- into variables, CA-CSPN can adaptively assemble results from different steps in the propagation. However, in practice, the computational complexity is too high to be deployed in real applications. To speed up the algorithm, RA-CSPN picks the optimal kernel size and the number of iterations for each pixel depending on the learned hyper-parameters. With the improvements, the method can dynamically allocate the context and computing resources required by each pixel. 

Unlike CSPN++, NLSPN~\citep{park2020non} is proposed with a two-stage network based on non-local neighbors instead of a fixed receptive field. In the first stage, a U-net is employed to produce a coarse depth, a confidence map, and non-local neighbors prediction with their raw affinities. Then, non-local spatial propagation computed by deformable convolutions~\citep{zhu2019deformable} is performed iteratively. In this manner, irrelevant features are avoided, while only relevant non-local neighborhood features are emphasized. Similar to NLSPN~\citep{park2020non}, DSPN~\citep{xu2020deformable} also conducts spatial propagation with deformable convolution, which can offer an adaptive receptive field according to the dependencies between pixels. 

Later, motivated by dynamic convolution~\citep{chen2020dynamic}, \cite{lin2022dynamic} develop a non-linear propagation model, named dynamic spatial propagation network (DySPN), where the affinity matrix is decoupled into parts depending on the distance. For neighborhoods with different distances, the affinity is assigned different weights to improve CSPN. Moreover, they propose a diffusion suppression operation, which adaptively stops iterations on specific pixels as directed by the attention matrix to avoid over-smoothing issues. According to different computational budgets, they design three variants of adaptive affinity matrix to achieve a balance between performance and complexity. 
 
\cite{liu2022graphcspn} integrate SPN and graph neural networks into one framework, allowing to learn both neighboring and long-range features. 
To adapt to practical applications, \cite{schuster2021ssgp,conti2023sparsity} propose novel approaches that can achieve good generalization.
\cite{Zhang2023CompletionFormer} adopt the non-local spatial propagation network for improving the depth quality after obtaining an initial densified depth map. As for the backbone that produces an initial depth, they propose a novel single-branch network, which leverages both the advantages of CNNs and transformers. In the network, they design a joint convolutional attention and transformer (JCAT) block that contains two paths -- the transformer layer and the convolutional attention layer -- and achieves more efficient performance than pure transformer-based methods. To learn the neighborhood affinity, \cite{lee2022multi} propose a network that leverages multi-scale and local features.
\revise{\citep{LRRU_ICCV_2023} propose a long-short range recurrent updating (LRRU) network that leverages an iterative scheme and employs fewer parameters to achieve state-of-the-art performance.}

\subsection{Auxiliary-based Methods}
As described in Sec.~\ref{aux_gdsr}, auxiliary learning employs several related tasks to improve the performance of the main task~\citep{zhang2021survey}. Many methods using auxiliary learning also emerge in GDC.
\cite{atapour2019complete} propose a multi-task framework, which allows conducting monocular depth estimation and sparse depth completion. This network is implemented as two cascaded sub-networks based on encoder-encoder architecture, with the first yielding a sparse depth map from a color image and the second estimating a dense depth map, taking the previous sparse depth as input. To obtain sharp depth boundaries in the final depth maps, they adopt adversarial training on both synthetic and real-world datasets. 

\begin{figure}[t]
   \centering
    \begin{tabular}{c}
		\includegraphics[width=0.9\linewidth]{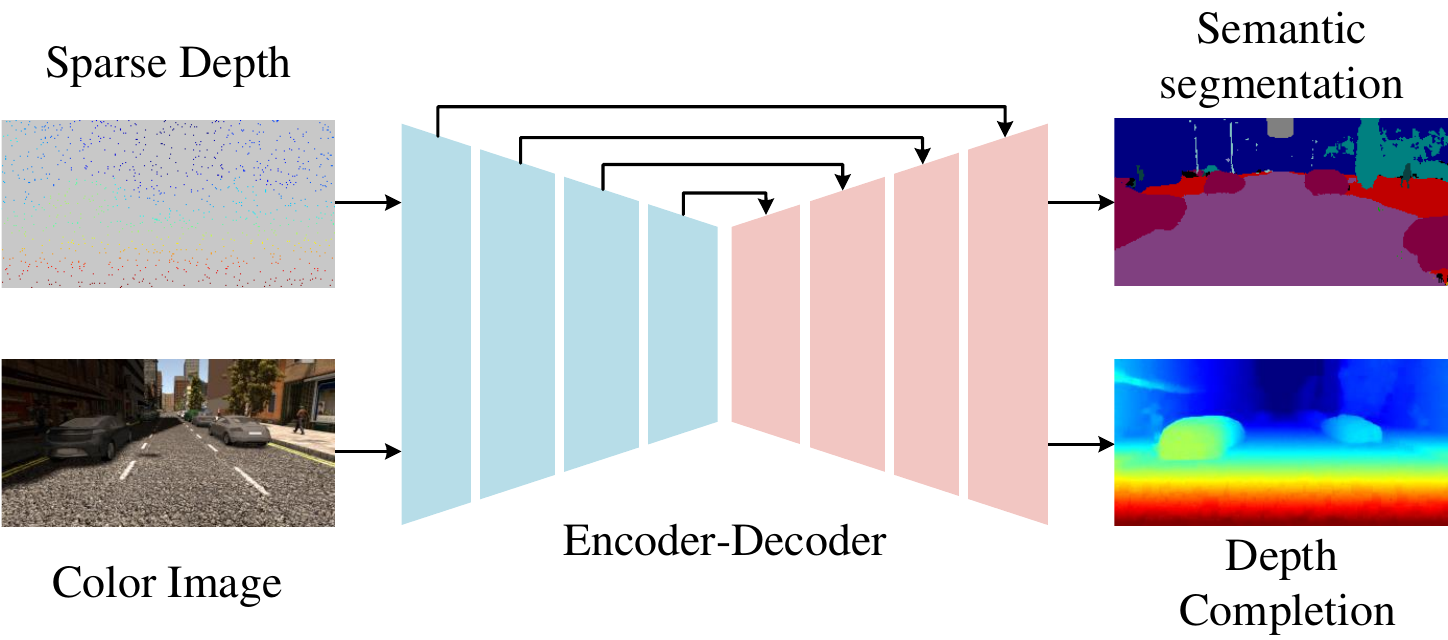} 
    \end{tabular}
    \caption{MNASnet framework~\citep{jaritz2018sparse} {belonging to auxiliary-based methods.} It can perform depth completion and semantic segmentation simultaneously with minor changes in the last layer. }
    \label{fig:sr_mnasnet}
\end{figure}

In addition to monocular depth estimation, semantic segmentation also contributes to the performance improvement of depth completion. \cite{jaritz2018sparse} perform multi-task learning, namely depth completion and semantic segmentation as shown in Fig. \ref{fig:sr_mnasnet}, by designing different heads at the end of a common network, e.g., NASNet~\citep{zoph2018learning}. When depth density varies, the method can still fuse features across modalities effectively. Unlike the previous method, which does not utilize results from the auxiliary task, \cite{zhang2021multitask} use semantic maps to improve the accuracy of depth completion. In order to integrate semantic segmentation and depth completion into a single framework, they build multi-task generative adversarial networks in combination with a semantic-guided smoothness loss, which is inspired by~\cite{zhao2020masked}. In addition to semantic segmentation and depth completion, edge detection, which serves as a bridge between the two tasks, is introduced into {the simultaneous semantic segmentation and depth completion multi-task network (SSDNet)}~\citep{zou2020simultaneous}. Based on the multi-task learning of semantic segmentation and depth completion, \cite{hirata2019real} utilize boundary class labeling to adjust the effects of adjacent pixels. 
In~\cite{liang2019multi}, the authors perform multiple perception tasks jointly, including 2D and 3D object detection, ground estimation and depth completion, to obtain better representations. Although this work is aimed at object detection, it demonstrates that the selected tasks are complementary, so helpful cues provided by the other tasks can also be applied to depth completion.
Motivated by \cite{lu2020depth}, \cite{ramesh2023siunet} design an auxiliary task to generate depth contours while outputting dense depth maps at the end of the network.

 Researchers leverage some further auxiliary information to enhance the precision of densified depth maps. Through a modified auto-encoder framework, \cite{zhang2018deep,qiu2019deeplidar} propose to estimate surface normal maps from color images to improve the depth accuracy. In~\cite{qiu2019deeplidar}, the surface normals are regarded as intermediate representations since they can result in higher errors in distance measures. Considering the limitation of surface normals, the dense depth map estimated by the color branch is also considered in the generation of the final dense depth. Instead of generating a surface normal from a color map, \cite{liu2022nnnet} take the normal map produced by sparse depth maps as an intermediate constraint during training.
\cite{xu2019depth} introduce a unified two-stage framework based on the assumption that natural scenes can be represented by piecewise planes. They build a diffusion module that incorporates the geometric constraints between depth and surface normals. This module is differentiable, allowing the diffusion conductance to be adjusted flexibly depending on the similarity in the guide.  
Inspired by \cite{van2019sparse,hu2021penet}, \cite{nazir2022semattnet} added a semantic-guided branch to highlight depth discontinuities. 

Instead of using surface normals as auxiliary information, some works utilize a point cloud to enhance model performance. In order to cope with unpredictable real-world environments, \cite{jeon2021abcd} propose a point-cloud-centric method based on the proposed attentive bilateral convolutional layer. This 3D convolution layer consists of four steps, i.e., splat, convolve, attention, and slice, which can directly perform convolution operations on point clouds and focus on capturing helpful features for GDC. 

Sometimes, depth completion as auxiliary information can also improve the performance of other tasks. For instance, \cite{carranza2022object,wu2022sparse} integrate depth completion networks into object detection models to achieve more accurate results.

\subsection{Fusion-focused Methods}
Similar to GDSR, fusion strategies are also crucial for GDC. Compared with straightforward fusion methods, such as concatenation~\citep{eldesokey2019confidence,zhu2022robust} or summation~\citep{shivakumar2019dfusenet,ryu2021scanline}, researchers design various complicated and effective methods.

\cite{lee2020deep} propose a multi-scale network that exchanges information with the attention mechanism at each encoder scale. 
\cite{yan2020revisiting} employ a similar strategy to fuse cross-modality features at multiple scales.
\cite{zhao2021adaptive} conduct multi-modal information fusion via the proposed symmetric gated fusion strategy consisting of two paths, each aiming to modulate the current modality features with adaptively extracted features from the other modality. Using this strategy, features can interact across modalities, thus allowing the model to obtain accurate depth estimates. 
\cite{liu2021fcfr} develop a novel coarse-to-fine framework that emphasizes the fusion of cross-modal information. It introduces a channel shuffle extraction operation to blend features from different modalities. Next, they propose an energy-based fusion operation, which selects feature values with higher regional energy to fuse the features effectively. 
\cite{ke2021mdanet} propose {a multi-modal deep aggregation network (MDANet)}, containing multiple connection and aggregation pathways to implement a deeper fusion of the two modalities.
\cite{liu2023mff} design a multi-feature channel shuffle extraction, which utilizes the channel shuffle operation to fuse features from color images and depth maps in the encoding stage. Then, a multi-level weighted combination leveraging multi-scale features is introduced into the decoding process to enhance the information fusion effectiveness further.

\begin{figure}[t]
   \centering
    \begin{tabular}{c}
		\includegraphics[width=0.95\linewidth]{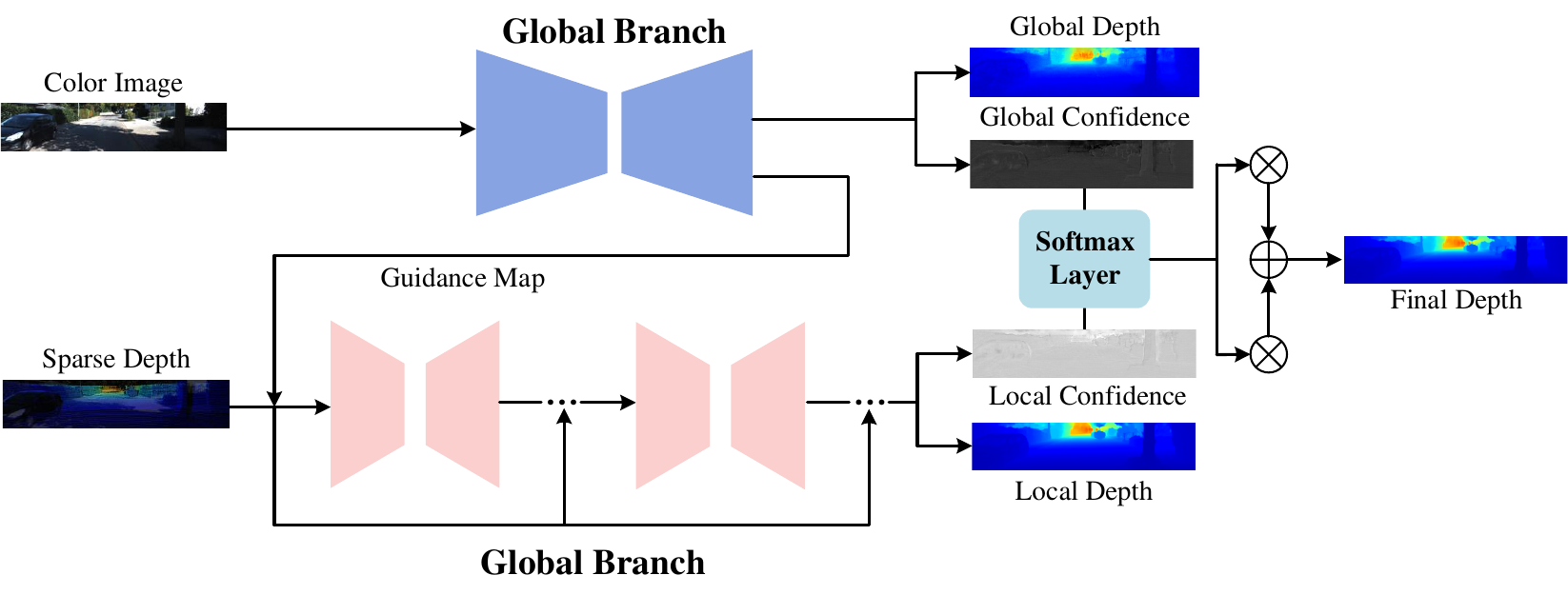} 
    \end{tabular}
    \caption{The framework of FusionNet~\citep{van2019sparse} {belonging to the category of fusion strategy.} This method fuses global and local features based on the confidence maps in a late fusion manner.}
    \label{fig:FusionNet}
\end{figure}

\subsection{Uncertainty Guidance} 
Most of the works surveyed so far do not take into account the reliability of the sparse depth when designing algorithms. Recently, two categories of uncertainty, i.e., aleatoric and epistemic uncertainty, have been considered in GDC with Bayesian deep learning. Aleatoric  
or stochastic uncertainty represents inherent randomness in each observation of the same experiment. In the task of GDC, aleatoric uncertainty refers to the random noise in raw LiDAR data and the distribution of sparse pixels. Epistemic or systematic uncertainty is often caused by models ignoring certain practical factors. Hence, the models for GDC are usually focused on handling aleatoric uncertainty.

\cite{eldesokey2018propagating,eldesokey2019confidence} introduce a novel framework with a proposed algebraically-constrained normalized convolution layer, where confidence can be computed and propagated to subsequent layers. Here, confidence is used as auxiliary information to improve the model performance. Furthermore, they propose a loss function by maximizing the final confidence to constrain the training. 
Later, \cite{eldesokey2020uncertainty} estimate the confidence of the input depth with the normalized convolution layer in an unsupervised manner, which can significantly improve the accuracy of depth prediction.  

As shown in Fig.~\ref{fig:FusionNet}, \cite{van2019sparse} estimate confidence maps in the global and local branches, respectively. Based on the learned confidence maps calculated by the softmax function in the last layer, the model can pay more attention to reliable pixels. Sharing a similar idea, \cite{qiu2019deeplidar} output confidence maps for occlusion handling and integrate depth predictions from color images and normal as well. \cite{xu2019depth} obtain confidence maps through a decoder to suppress noise propagation in depth maps.

\cite{zhu2022robust} provide a novel uncertainty formulation with Jeffrey's prior, which can improve the robustness of the model to noise or invalid pixels. In the first stage, this proposed multi-scale joint prediction model outputs coarse depth and uncertainty maps, which are then sent to the uncertainty attention residual learning network for depth refinement. Through their architecture and loss function, outliers and uneven distribution of pixels are well handled. Additionally, \cite{peng2022pixelwise} attempt to employ the proposed uncertainty-aware sampling algorithm to identify the reliability of pixels, which is beneficial for the subsequent adaptive prediction of the depth range for each pixel.

\begin{figure}[t]
   \centering
    \begin{tabular}{c}
		\includegraphics[width=0.9\linewidth]{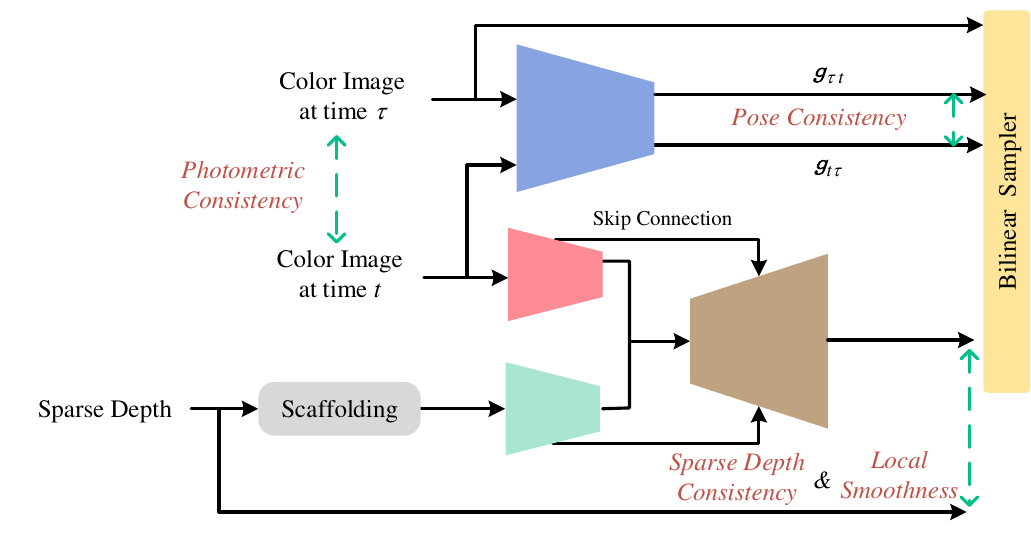} 
    \end{tabular}
    \caption{VIO~\citep{wong2020unsupervised} framework {belonging to unsupervised methods.} Combining the four losses of photometric consistency, sparse depth consistency, pose consistency and local smoothness enables training the network in a fully unsupervised manner.}
    \label{fig:VIO}
\end{figure}

\begin{table*}[htbp] 
    \centering
    \scalebox{0.6}{
    \begin{tabular}{cllcccc}
    \hline
    Paradigm        & Method      & Reference           & \multicolumn{4}{c}{Key Idea} \\ \hline
 
    \multirow{26}{*}{\rotatebox{90}{\textbf{Dual-branch}}}  
    
    &  MNASNet~\citep{jaritz2018sparse}    & 3DV-2018        & \multicolumn{4}{l}{\begin{tabular}[l]{@{}l@{}} Leverage fused features from dual streams to yield\ the final dense estimation \\via a decoder.\end{tabular}} \\ 
    
    &  \multirow{2}{*}{GuideNet~\citep{tang2020learning}}    &  \multirow{2}{*}{TIP-2020}       & \multicolumn{4}{l}{\multirow{2}{*}{\begin{tabular}[l]{@{}l@{}} Propose a hourglass network for depth completion.\end{tabular}}} \\ 
    &   &   &   \\ 

    &  \multirow{2}{*}{KernelNet ~\citep{liu2021learning}}    &  \multirow{2}{*}{TIP-2021}       & \multicolumn{4}{l}{\multirow{2}{*}{\begin{tabular}[l]{@{}l@{}} Present a two-stage model where depth interpolation and refinement are\\ performed sequentially.\end{tabular}}} \\ 
    &   &   &   \\  
    
    &  \multirow{2}{*}{RigNet~\citep{yan2022rignet}}   & \multirow{2}{*}{ECCV-2022}        & \multicolumn{4}{l}{\multirow{2}{*}{\begin{tabular}[l]{@{}l@{}} Exploit a repetitive design based on the hourglass network.\end{tabular}}} \\ 
    &   &   &   \\  
    
    &  {MSG-CHN~\citep{li2020multi}}    &  {WACV-2020}       & \multicolumn{4}{l}{\multirow{2}{*}{\begin{tabular}[l]{@{}l@{}} Introduce the cascaded architecture into the hourglass network.\end{tabular}}} \\ 

    &  CDCNet ~\citep{Fan_2022_BMVC}   &  BMVC-2022     &  \\ 
    
    &  \multirow{2}{*}{Penet~\citep{hu2021penet}}  &  \multirow{2}{*}{ICRA-2021}        & \multicolumn{4}{l}{\multirow{2}{*}{\begin{tabular}[l]{@{}l@{}} Enhance the accuracy of depth prediction in a coarse-to-fine manner.\end{tabular}}} \\ 
    &   &   &   \\  
    
    &  FusionNet~\citep{van2019sparse}   &  MVA-2019       & \multicolumn{4}{l}{\begin{tabular}[l]{@{}l@{}} Exploit confidence maps in the global and local branches to achieve a guided-\\base method.\end{tabular}} \\ 
    
    &  \multirow{2}{*}{DenseLiDAR~\citep{gu2021denselidar}}   &  \multirow{2}{*}{RAL-2021}       & \multicolumn{4}{l}{\multirow{2}{*}{\begin{tabular}[l]{@{}l@{}} Use a dense pseudo-depth map as a coarse reference for the final prediction.\end{tabular}}} \\ 
    &   &   &   \\ 
    
    &  \multirow{2}{*}{SDDC~\citep{xu2023real}}   &  \multirow{2}{*}{Visual Computer-2023}       & \multicolumn{4}{l}{\multirow{2}{*}{\begin{tabular}[l]{@{}l@{}} Propose the semantic-guided branch to achieve better prediction results.\end{tabular}}} \\ 
    &   &   &   \\  
    
    &  QNN~\citep{jiang2022low}  &  CVPR-2022        & \multicolumn{4}{l}{\begin{tabular}[l]{@{}l@{}} Adopt a dual-branch structure to achieve an efficient solution with low power\\ consumption.\end{tabular}} \\ 
    
    &  ACMNet~\citep{zhao2021adaptive}   & TIP-2021        & \multicolumn{4}{l}{\begin{tabular}[l]{@{}l@{}} Conduct multi-modal information fusion via the proposed symmetric gated \\fusion strategy.\end{tabular}} \\

    &  \multirow{2}{*}{\revise{PointDC~\citep{yu2023aggregating}}}   &  \multirow{2}{*}{\revise{ICCV-2023}}       & \multicolumn{4}{l}{\multirow{2}{*}{\begin{tabular}[l]{@{}l@{}} \revise{Propose a dual-branch network to extract 2d and sparse 3D feature point} \\ \revise{cloud for depth completion.}\end{tabular}}} \\ 
    &   &   &   \\  
    
    &  \multirow{2}{*}{\revise{AGGNet~\citep{chen2023agg}}}   &  \multirow{2}{*}{\revise{ICCV-2023}}       & \multicolumn{4}{l}{\multirow{2}{*}{\begin{tabular}[l]{@{}l@{}} \revise{Design a dual-branch autoencoder with attention mechenism to obtain dense} \\ \revise{depth maps.}\end{tabular}}} \\ 
    &   &   &   \\      \hline
    
    \multirow{22}{*}{\rotatebox{90}{\textbf{Affinity-based}}}  
    
    &  SPN~\citep{liu2017learning}    &  NeurIPS-2017       & \multicolumn{4}{l}{\begin{tabular}[l]{@{}l@{}} Exploit the spatial propagation network to learning affinity matrix for depth \\completion.\end{tabular}} \\ 

    &  CSPN~\citep{cheng2018depth}    &  ECCV-2018       & \multicolumn{4}{l}{\begin{tabular}[l]{@{}l@{}} Formulate the task as anisotropic diffusion filtering, and implement it through a\\ deep network.\end{tabular}} \\ 

    &  \multirow{2}{*}{CSPN++~\citep{cheng2020cspn++}}   &  \multirow{2}{*}{AAAI-2020}       & \multicolumn{4}{l}{\multirow{2}{*}{\begin{tabular}[l]{@{}l@{}} Propose a new structure including context-aware and resource-aware CSPNs.\end{tabular}}} \\ 
    &   &   & \\ 

    &  NLSPN~\citep{park2020non}     &  ECCV-2020      & \multicolumn{4}{l}{\begin{tabular}[l]{@{}l@{}} Propose a two-stage network that is based on non-local neighbors instead of a \\fixed receptive field.\end{tabular}} \\ 

    &  \multirow{2}{*}{DSPN~\citep{xu2020deformable}}    &  \multirow{2}{*}{ICIP-2020}       & \multicolumn{4}{l}{\multirow{2}{*}{\begin{tabular}[l]{@{}l@{}} Conduct spatial propagation with deformable convolution.\end{tabular}}} \\ 
    &   &   &   \\  

    &  DySPN~\citep{lin2022dynamic}   &  AAAI-2022       & \multicolumn{4}{l}{\begin{tabular}[l]{@{}l@{}} Develop a non-linear propagation model, where the affinity matrix is decoupled\\ into parts depending on the distance.\end{tabular}} \\ 

    &  GraphCSPN~\citep{liu2022graphcspn}    &  ECCV-2022       & \multicolumn{4}{l}{\begin{tabular}[l]{@{}l@{}} Integrate SPN and graph neural networks to learn both neighboring and long-\\range features.\end{tabular}} \\ 

    &  Ssgp~\citep{schuster2021ssgp}     &  WACV-2021     & \multicolumn{4}{l}{\begin{tabular}[l]{@{}l@{}} Propose a unified architecture to perform sparse-to-dense interpolation in \\different domains.\end{tabular}} \\ 

    &  SpAgNet~\citep{conti2023sparsity}  &  WACV-2023       & \multicolumn{4}{l}{\begin{tabular}[l]{@{}l@{}} Propose a depth completion method that is independent of their distribution and \\density.\end{tabular}} \\

    &  CompletionFormer~\citep{Zhang2023CompletionFormer}    &  CVPR-2023      & \multicolumn{4}{l}{\begin{tabular}[l]{@{}l@{}} Leverage the non-local spatial propagation network to improve the depth \\quality after obtaining initial depth with convolutions and transformers.\end{tabular}} \\ 

    & \multirow{2}{*}{DenseLConv~\citep{lee2022multi}}   &  \multirow{2}{*}{IROS-2022}       & \multicolumn{4}{l}{\multirow{2}{*}{\begin{tabular}[l]{@{}l@{}} Propose a depth completion method that is independent of their distribution \\and density.\end{tabular}}} \\ 
    &   &   &   \\

    & \multirow{2}{*}{\revise{LRRU~\citep{LRRU_ICCV_2023}}} &  \multirow{2}{*}{\revise{ICCV-2023}} &  \multicolumn{4}{l}{\multirow{2}{*}{\begin{tabular}[l]{@{}l@{}} \revise{Propose a lightweight network to learn spatially-variant kernels and perform} \\ \revise{updates iteratively.} \end{tabular}}} \\    
    &   &   &   \\  \hline

    \multirow{20}{*}{\rotatebox{90}{\textbf{Auxiliary-based}}}

    &  DCMDE~\citep{atapour2019complete}    &  3DV-2019       & \multicolumn{4}{l}{\begin{tabular}[l]{@{}l@{}} Propose a multi-task framework conducting monocular depth estimation and\\ sparse depth completion.\end{tabular}} \\ 

    &  Multitask GANs~\citep{zhang2021multitask}   &  TNNLS-2021       & \multicolumn{4}{l}{\begin{tabular}[l]{@{}l@{}} Present a multi-task network using semantic images to improve the accuracy\\ of depth completion.\end{tabular}} \\ 

    &  SSDNet ~\citep{zou2020simultaneous}    &  Sensors-2020       & \multicolumn{4}{l}{\begin{tabular}[l]{@{}l@{}} Introduce edge detection to serve as a bridge between semantic segmentation\\ and depth completion.\end{tabular}} \\

    &  MMF ~\citep{liang2019multi}   &  CVPR-2019       & \multicolumn{4}{l}{\begin{tabular}[l]{@{}l@{}} Combine 2D and 3D object detection, ground estimation, and depth completion \\to obtain better representations.\end{tabular}} \\ 

    &  DPS ~\citep{hirata2019real}  &  RAL-2019       & \multicolumn{4}{l}{\begin{tabular}[l]{@{}l@{}} Present a method for estimating a dense depth from a sparse LIDAR point cloud \\and an image sequence.\end{tabular}} \\ 

    &  SIUNet ~\citep{ramesh2023siunet}  &  WACV-2023        & \multicolumn{4}{l}{\begin{tabular}[l]{@{}l@{}} Propose a network regarding depth contours as auxiliary information to obtain\\ dense depth maps.\end{tabular}} \\ 

    &  DDC ~\citep{zhang2018deep}  &  CVPR-2018       & \multicolumn{4}{l}{\multirow{3}{*}{\begin{tabular}[l]{@{}l@{}} Propose to estimate surface normal maps from color images to improve the\\ depth accuracy.\end{tabular}}} \\ 

    &  Deeplidar ~\citep{qiu2019deeplidar}   &  CVPR-2019     &  \\ 
    &  NNNet ~\citep{liu2022nnnet}  &  IEEE Access-2022       &  \\ 

    &  DNC ~\citep{xu2019depth}   &  ICCV-2019      & \multicolumn{4}{l}{\begin{tabular}[l]{@{}l@{}} Confidence maps are obtained through a decoder to suppress noise propagation \\in depth maps.\end{tabular}} \\ 

    &  SemAttNet ~\citep{nazir2022semattnet} &  IEEE Access-2022      & \multicolumn{4}{l}{\begin{tabular}[l]{@{}l@{}} Add a semantic-guided branch for depth completion to highlight depth \\discontinuities.\end{tabular}} \\ 

    &  ABCD ~\citep{jeon2021abcd}   &  RAL-2021      & \multicolumn{4}{l}{\begin{tabular}[l]{@{}l@{}} Propose a point-cloud-centric method that is based on the proposed attentive\\ bilateral convolutional layer.\end{tabular}} \\    \hline
    \end{tabular}}
    \caption{List of Guided Depth Completion methods, divided according to the first three outlined categories.}
    \label{gdc_list_1}
\end{table*}

\begin{table*}[t] 
    \centering
    \scalebox{0.52}{
    \begin{tabular}{cllcccc}
    \hline
    Paradigm        & Method      & Reference           & \multicolumn{4}{c}{Key Idea} \\ \hline

    \multirow{7}{*}{\rotatebox{90}{\textbf{Fusion-focused}}}

    &  CGM ~\citep{liu2021learning}  &  IEEE Access-2020       & \multicolumn{4}{l}{\begin{tabular}[l]{@{}l@{}} Present a two-stage model where depth interpolation and refinement are \\performed sequentially.\end{tabular}} \\ 

    &  \multirow{2}{*}{RSIC ~\citep{yan2020revisiting}}  &  \multirow{2}{*}{IEEE Access-2020}       & \multicolumn{4}{l}{\multirow{2}{*}{\begin{tabular}[l]{@{}l@{}} Employ a similar strategy to fuse cross-modality features at multiple scales.\end{tabular}}} \\ 
    &   &   &   \\ 

    &  FCFR-Net ~\citep{liu2021fcfr}   &  AAAI-2021     & \multicolumn{4}{l}{\begin{tabular}[l]{@{}l@{}} Develop a novel coarse-to-fine framework that emphasizes the fusion of\\ cross-modal information.\end{tabular}} \\

    &  MDANet ~\citep{ke2021mdanet}   &  ICRA-2021     & \multicolumn{4}{l}{\begin{tabular}[l]{@{}l@{}} Propose a multi-modal deep aggregation block that consists of multiple \\ connection and aggregation pathways for deeper fusion.\end{tabular}} \\ 

    &  MFF-Net ~\citep{liu2023mff}  &  RAL-2023       & \multicolumn{4}{l}{\begin{tabular}[l]{@{}l@{}} Propose a multi-feature channel shuffle extraction and a decoding process with\\ multi-scale features.\end{tabular}} \\ \hline

    \multirow{10}{*}{\rotatebox{90}{\textbf{Uncertainty-guided}}}

    &  MS-Net ~\citep{eldesokey2019confidence}  &  TPAMI-2019      & \multicolumn{4}{l}{\begin{tabular}[l]{@{}l@{}} Introduce a novel framework with a proposed algebraically-constrained \\normalized convolution layer.\end{tabular}} \\ 
     
    &  pNCNN ~\citep{eldesokey2020uncertainty}   &  CVPR-2020      & \multicolumn{4}{l}{\begin{tabular}[l]{@{}l@{}} Estimate the confidence of the input depth with the normalized convolution \\layer in an unsupervised manner.\end{tabular}} \\ 

    &  MJPM ~\citep{zhu2022robust}  &  AAAI-2022     & \multicolumn{4}{l}{\begin{tabular}[l]{@{}l@{}}  Provide Jeffrey's prior uncertainty formulation that can increase the model's \\resistance to noise and invalid pixel input.\end{tabular}} \\ 

    &  \multirow{2}{*}{FusionNet~\citep{wong2021learning}}   &  \multirow{2}{*}{RAL-2021}       & \multicolumn{4}{l}{\multirow{2}{*}{\begin{tabular}[l]{@{}l@{}} Leverage synthetic datasets to alleviate the lack of real image pairs.\end{tabular}}} \\ 
    &   &   &   \\  

    &  PADNet ~\citep{peng2022pixelwise}  &  ACM MM-2022      & \multicolumn{4}{l}{\begin{tabular}[l]{@{}l@{}} Employ the proposed uncertainty-aware sampling algorithm to identify the\\ reliability of pixels, which is beneficial for prediction of the depth range.\end{tabular}} \\

    \hline

    \multirow{20}{*}{\rotatebox{90}{\textbf{Unsupervised}}} 
    
    &  SfMLearner~\citep{zhou2017unsupervised}    &   CVPR-2017      & \multicolumn{4}{l}{\begin{tabular}[l]{@{}l@{}} Single-view depth prediction and multi-view camera pose estimation from two \\independent networks.\end{tabular}} \\ 
    
    &  \multirow{2}{*}{SSDC~\citep{ma2019self}}   &  \multirow{2}{*}{ICRA-2019}       & \multicolumn{4}{l}{\multirow{2}{*}{\begin{tabular}[l]{@{}l@{}} Propose learning a mapping from the sparse input to the dense prediction.\end{tabular}}} \\ 
    &       &       &   \\  
    
    &  lsf~\citep{qu2020depth}   & WACV-2020        & \multicolumn{4}{l}{\begin{tabular}[l]{@{}l@{}} Propose a least squares fitting model for the self-supervised learning benchmark\\ network.\end{tabular}} \\ 
    
    &  VIO~\citep{wong2020unsupervised}   & ICRA-2020        & \multicolumn{4}{l}{\begin{tabular}[l]{@{}l@{}} Use a piecewise planar scaffolding of a scene as supervision to achieve self-\\supervised learning.\end{tabular}} \\ 
    
    &  \multirow{2}{*}{KBNet~\citep{wong2021unsupervised}}   &  \multirow{2}{*}{ICCV-2021}      & \multicolumn{4}{l}{\multirow{2}{*}{\begin{tabular}[l]{@{}l@{}} Exploit the proposed calibrated backprojection layers to improve the baselines.\end{tabular}}} \\ 
    &       &       &   \\  
    
    &  \multirow{2}{*}{Monitored Distillation~\citep{liu2022monitored}}   & \multirow{2}{*}{ECCV-2022}  & \multicolumn{4}{l}{\multirow{2}{*}{\begin{tabular}[l]{@{}l@{}} Introduce knowledge distillation into depth completion.\end{tabular}}} \\ 
    &       &       &   \\  
    
    &  \multirow{2}{*}{CostDCNet~\citep{kam2022costdcnet}}   &  \multirow{2}{*}{ECCV-2022}       & \multicolumn{4}{l}{\multirow{2}{*}{\begin{tabular}[l]{@{}l@{}} Propose a two-branch network based on the cost volume-based depth estimation.\end{tabular}}} \\ 
    &       &       &   \\  
    
    &  \multirow{2}{*}{Struct-MDC~\citep{jeon2022struct}}   &  \multirow{2}{*}{RAL-2022}       & \multicolumn{4}{l}{\multirow{2}{*}{\begin{tabular}[l]{@{}l@{}} First perform mesh depth construction by leveraging point and line features.\end{tabular}}} \\ 
    &       &       &   \\  
    
    &  \multirow{2}{*}{DDP~\citep{yang2019dense}}   &  \multirow{2}{*}{CVPR-2019}      & \multicolumn{4}{l}{\multirow{2}{*}{\begin{tabular}[l]{@{}l@{}} Develop a conditional prior network to calculate a posterior probability over the\\ depth of each pixel.\end{tabular}}} \\ 
    &       &       &   \\  
    
    &  \multirow{2}{*}{AGG-CVCNet~\citep{xu2022self}}   &  \multirow{2}{*}{ACM MM-2022}       & \multicolumn{4}{l}{\multirow{2}{*}{\begin{tabular}[l]{@{}l@{}} Propose a novel adjacent geometry guided training loss.\end{tabular}}} \\ 
    &   &   &   \\   
    
    &  MSC~\citep{zhang2022self}    & Remote Sensing-2022      & \multicolumn{4}{l}{\multirow{2}{*}{\begin{tabular}[l]{@{}l@{}} Introduce multi-modal spatio-temporal consistency constraints to train models.\end{tabular}}} \\
    
    &  LeoVR~\citep{li2022motion}    &  MobiSys-2022       &  \\ \hline

    \end{tabular}}
    \caption{List of Guided Depth Completion methods, divided according to the last three outlined categories.}
    \label{gdc_list_2}
\end{table*}

\subsection{Unsupervised Methods}
In practice, dense ground truth depth maps are hard to obtain due to the limitations of LiDAR imaging principles. Even a high-end LiDAR can only generate at most $30\%$ valid pixels of the whole depth map~\citep{uhrig2017sparsity}. Thus, some methods attempt to address the problem with self-supervised learning, among which the representative classes build supervisory signals using monocular sequences or synchronized stereo pairs. 

As depicted in Fig.~\ref{fig:VIO}, \citep{zhou2017unsupervised} design a self-supervised framework in which only color and depth sequences must be provided instead of sparse/dense image pairs. The method produces single-view depth prediction and multi-view camera pose estimation from two independent networks. This principle is brought to GDC by
\citep{ma2019self} introducing three loss functions, i.e., sparse depth loss, photometric loss, and smoothness loss. Under this framework, the performance even outperforms some methods with semi-dense annotations. 
To increase the robustness to noise and outliers, \citep{qu2020depth} substituted the proposed least squares fitting model for the last convolutional layer of the self-supervised learning benchmark network~\citep{godard2019digging}.
\cite{wong2020unsupervised} propose to use a piecewise planar scaffolding of a scene as supervision to implement self-supervised learning. Using a similar structure, \cite{wong2021unsupervised} leverage the proposed calibrated back-projection layers to improve the baselines, while \cite{liu2022monitored} introduce knowledge distillation into depth completion. 
\cite{kam2022costdcnet} propose a two-branch network inspired by cost volume-based depth estimation, which can obtain informative 2D and 3D representations from cross-modality input with the lightweight structure design. In \cite{jeon2022struct}, they first perform mesh depth construction by leveraging point and line features, which, together with the color image, are fed into a mesh depth refinement module. Finally, the depth prediction can be obtained through {the calibrated backprojection network (KBNet)}~\citep{wong2021unsupervised}.
While most networks are trained on real datasets, \cite{wong2021learning,lopez2020project} train their models on synthetic datasets, alleviating the lack of real image pairs with dense labels.

In unsupervised learning, stereo pairs also play an important role. Among the self-supervised stereo methods, most use photometric consistency~\citep{godard2017unsupervised,yang2019dense,shivakumar2019dfusenet} to infer geometry and obtain promising results.    
In contrast, \cite{xu2022self} propose a novel adjacent geometry guided training loss to confine depth maps of low-confident regions by high-confident labels.  \cite{zhang2022self,li2022motion} introduce multi-modal spatiotemporal consistency constraints to train models, which enables the model to better adapt to real-world environments, such as dark and low-texture objects.

To conclude this section, Tabs. \ref{gdc_list_1} and \ref{gdc_list_2} collect the methods discussed so far, divided into six categories. For each method, the venue and year are reported, together with a short description of the key idea behind it.

\section{Benchmark Datasets and objective functions} \label{datasets}

We introduce existing datasets relevant to our survey and the most common loss functions used to train the methods we surveyed before.

\subsection{Datasets}
This section introduces public datasets for RGB guided ToF imaging; they are divided into low-resolution and sparse depth according to the image characteristics collected by ToF cameras.

\begin{table*}[t] \scriptsize
	\centering
        \scalebox{0.6}{
	\begin{tabular}{lccccccc}
	\toprule
	   & \textbf{Dataset} & \textbf{Ref.} & \textbf{Year} & \textbf{Sensor Name} & \textbf{Capture condition}  & \textbf{Modalities} & \textbf{Images}    
        \\ \midrule
        \multirow{14}{*}{\rotatebox[origin=b]{90}{\textbf{LR Depth}}} & ToFMark &~\cite{ferstl2013image} & 2013  & PMD Nano & Indoor &Color, Depth  & 3 Images     \\
	& Lu &~\cite{lu2014depth} & 2014 & ASUS Xiton Pro & Indoor & Color, Depth & 6    \\
        & SUN RGBD &~\cite{song2015sun} & 2015  & Kinect v2  & Indoor  & Color, Depth  & 2860        \\ 
        & DIML &~\cite{cho2021deep} & 2021 & \tabincell{c}{Kinect v2, \\ ZED Stereo}   & \tabincell{c}{Indoor, \\  Outdoor} & Color, Depth  & $>$200 Scenes   \\
        & RGBDD &~\cite{he2021towards} & 2021  & \tabincell{c}{ Huawei P30 Pro, \\ Helios}  & \tabincell{c}{Indoor, \\  Outdoor} & Color, Depth  & 4811 Images    \\
        & \multirow{3}{*}{Middlebury} & \multirow{3}{*}{\begin{tabular}[c]{@{}c@{}}~\cite{scharstein2003high} \\~\cite{hirschmuller2007evaluation} \\~\cite{scharstein2014high} \end{tabular}} & \multirow{3}{*}{\begin{tabular}[c]{@{}c@{}}2003\\2007\\2014 \end{tabular}}   & \multirow{3}{*}{\begin{tabular}[c]{@{}c@{}}Structured Light\\Structured Light\\Stereo Camera \end{tabular}}   & \multirow{3}{*}{Indoor}    & \multirow{3}{*}{Color,Depth}   & \multirow{3}{*}{32 Images}     \\
        & &  &  &  &    &  \\
        & &  &  &  &    &  \\
	& NYUv2 &~\cite{silberman2012indoor} & 2012 & Kinect V1 & Indoor  & Color, Depth & 1449 Images      \\ 
        & MPI Sintel Depth &~\cite{butler2012naturalistic} & 2012 & Blender & \tabincell{c}{Indoor, \\  Outdoor} & \tabincell{c}{Color, Depth \\ Optical Flow}  & 1064 Images    \\
        & ToF-FT3D &~\cite{wang2022self} & 2022  & synthetic  & --  & Color, Depth  & 6250 Views     \\      \hline
        \multirow{14}{*}{\rotatebox[origin=b]{90}{\textbf{Sparse Depth}}} & KITTI&~\cite{uhrig2017sparsity}	& 2017 & Velodyne, Stereo Camera & Outdoor  & Color, Depth  &  94k Frames   \\
        & DenseLivox&~\cite{yu2021grayscale}	& 2021  & \tabincell{c}{Livox LiDAR, \\ RealSense d435i}   & \tabincell{c}{Indoor, \\  Outdoor}  & Color, Depth  &  6 Scenes       \\
        & KITTI-360&~\cite{liao2022kitti} & 2022  & Velodyne, Stereo Camera  & Outdoor  & \tabincell{c}{Color, Depth, \\ GPS, IMU}  & 150 Scenes      \\
        & Gibson&~\cite{xia2018gibson}	& 2018  & \tabincell{c}{NaVis, Matterport Camera, \\DotProduct}   & Outdoor  & Color, Depth  & 572 Scenes      \\ 
        & Virtual KITTI	&~\cite{gaidon2016virtual}  & 2016  & Synthetic &Outdoor  & Color,Depth  & 50 Videos     \\  
        & Leddar Pixset	&~\cite{deziel2021pixset}  & 2021  & Leddar Pixell LiDAR & Outdoor  & \tabincell{c}{Color, Depth \\IMU, Radar}   & 29k Frames     \\ 
        & Waymo Perception	&~\cite{sun2020scalability}  & 2020  & LiDAR & Outdoor  & Color, Depth  & 1150 Scenes     \\
        & DDAD	&~\cite{guizilini20203d}  & 2020  & Luminar-H2 LiDAR & Outdoor  & Color, Depth  & 150 Scenes     \\
        & Near-Collision Set &~\cite{manglik2019forecasting}  & 2019  & \tabincell{c}{LiDAR (N/A), \\Stereo Camera} & Indoor  & Color, Depth  & 13658 Sequences      \\
        \bottomrule
        \end{tabular}}
	\caption{RGB guided ToF Datasets, divided according to the depth map properties.}
	\label{datasets_list}
\end{table*}

\subsubsection{Low-resolution depth}
Here, we mainly introduce datasets composed of low-resolution images collected by RGB guided ToF cameras, which can be employed for depth super-resolution and depth completion through certain sampling strategies. We also introduce datasets that are commonly used in this field, such as Middlebury~\citep{scharstein2003high,hirschmuller2007evaluation,scharstein2007learning,scharstein2014high}, NYU v2~\citep{silberman2012indoor}, MPI Sintel Depth~\citep{butler2012naturalistic} and ToF-FT3D~\citep{wang2022self}, although being collected by other sensors or synthesized.

\textbf{ToFMark~\citep{ferstl2013image}:} This dataset captures 3 real-world scenes, i.e., Books, Shark and Devil, by PMD Nano ToF camera. 

\textbf{LU~\citep{lu2014depth}:} This dataset consists of 6 RGBD images captured by the ASUS Xtion Pro camera, which usually serves as validation or test set. 

\textbf{SUN RGBD~\citep{song2015sun}:} SUN RGBD dataset contains 10\,335 indoor image pairs acquired by four different sensors. This dataset is often applied to scene understanding tasks. Additionally, this dataset can be used to investigate cross-sensor biases.

\textbf{DIML~\citep{cho2021deep}:} DIML is a large-scale RGBD database containing 2M RGBD frames. In order to obtain high-precision depth maps, they use Kinect v2 to capture indoor scenes and ZED stereo camera to capture outdoor scenes. In outdoor scenes, confidence maps of disparity are also provided.

\textbf{RGBDD~\citep{he2021towards}:} RGBDD establishes the first depth map SR dataset, which can reflect the correspondence between real LR and HR depth maps. This dataset consists of 4811 images capturing various scenes, e.g., human body, stuffed dolls, toys and plants. Notably, LR and HR depth maps are collected by Huawei P30 Pro and Helios ToF camera, respectively.

\textbf{Middlebury~\citep{scharstein2003high,hirschmuller2007evaluation,scharstein2007learning,scharstein2014high}:} The Middlebury dataset is a widely used dataset in the field of GDSR, and consists of four sub-datasets from different years. Middlebury 2003~\citep{scharstein2003high}, Middlebury 2005~\citep{hirschmuller2007evaluation}, Middlebury 2006~\citep{scharstein2007learning}, and Middlebury 2012~\citep{scharstein2014high} are acquired by the stereo camera under indoor condition. These datasets use stereo image pairs to generate depth maps. 

\textbf{NYUv2~\citep{silberman2012indoor}:} This dataset uses Microsoft Kinect v1 camera to capture 1449 images. Most previous works take the first 1000 RGBD images as the training set, and the rest of the 449 images as the test set. \cite{de2022learning,metzger2022guided} randomly selected 849 images for training, 300 for validation, and 300 for evaluation.

\textbf{MPI Sintel Depth~\citep{butler2012naturalistic}:} This dataset originates from an animated short film for flow evaluation. It provides synthetic video sequences, including 35 naturalistic scenes, for which depth maps are provided. 

\textbf{ToF-FT3D~\citep{wang2022self}:} ToF-FT3D, also known as ToF-FlyingThings3D, is a synthetic dataset which captures 6250 different views using Blender. Similar to the FlyingThings3D~\citep{mayer2016large} for optical flow estimation, it consists of objects that fly along randomized 3D trajectories. Besides, this dataset provides ToF amplitude, ToF depth and RGB images with a resolution of $640\times 480$.

\subsubsection{Sparse depth}
In practical applications, sparse depth maps are usually produced by LiDAR and only contain about $4\%$ valid pixels~\citep{uhrig2017sparsity}. Under some extreme conditions, even less than $1\%$ of the pixels are effectively measured~\citep{guizilini2020robust,guizilini2021sparse}. Although sparse depth maps can be obtained from dense depth maps with sampling strategies, here we focus on datasets collected by LiDAR.

\textbf{KITTI~\citep{uhrig2017sparsity}:} KITTI is a dataset initially gathered by the Karlsruhe Institute of Technology (KIT) and Toyota Technological Institute at Chicago (TTI-C). As one of the most important datasets in the field of autonomous driving, it captures more than 93K depth maps with corresponding raw LiDAR scans and RGB images. The LiDAR and two color cameras used for data acquisition are Velodyne HDL-64E and Point Grey Flea 2 (FL2-14S3C-C), respectively. Based on the KITTI raw dataset, \cite{uhrig2017sparsity} remove noise and artifacts in the scenes and aggregates multiple LiDAR scans over time to obtain denser ground truth depths, making this dataset suitable for GDC.
There are 86\,000, 7\,000, and 1\,000 image pairs for training, validation, and evaluation, respectively. 

\textbf{Virtual KITTI~\citep{gaidon2016virtual}:} This dataset is a photo-realistic synthetic video dataset that contains 50 videos with a total of 21\,260 frames. Scenes in the dataset are generated with the Unity game engine and a real-to-virtual cloning method. It can be applied to a variety of tasks, such as object detection, object tracking, instance semantic segmentation, and more.

\textbf{KITTI-360~\citep{liao2022kitti}:} KITTI-360 is a newly collected large-scale dataset with rich sensory information and full annotations. This dataset captures over 320k images and 100k laser scans, including annotated static and dynamic 3D scene elements. 

\textbf{Gibson~\citep{xia2018gibson}:} Gibson contains 572 full buildings with 1447 floors covering a total area of 211k $m^2$. In addition to depth maps and the corresponding color maps, it provides normal maps and semantic object annotations. 

\textbf{DenseLivox~\citep{yu2021grayscale}:} This dataset collects the images with a Livox Horizon LiDAR and Intel RalSense D435i camera. It contains dense, accurate depth as ground truth, with up to $88.3\%$ valid pixel coverage.

\textbf{Leddar Pixset~\citep{deziel2021pixset}:} Leddar Pixset is the first full-waveform flash LiDAR dataset for autonomous driving. It contains 29 000 frames in 97 sequences of various environments, weather conditions, and periods, annotated with more than 1.3 million 3D bounding boxes.

\textbf{Waymo Perception~\citep{sun2020scalability}:} This dataset is collected by five LiDAR sensors and five high-resolution pinhole cameras, all synchronized and calibrated. It captures 1150 scenes, including urban and suburban areas, each lasting 20 seconds. Typically, 1000 scenes are used for training and 150 for testing.

\textbf{DDAD~\citep{guizilini20203d}:} DDAD is a new benchmark dataset for  autonomous driving from Toyota Research Institute, which consists of monocular videos and accurate ground-truth depth. It has a long measurable distance, which can reach up to 250m. There are 150 scenes with 12 650 individual image pairs in the training set, 50 scenes with 3 950 image pairs in the validation set, and 3\,080 images in the test set.

\textbf{Near-Collision Set~\citep{manglik2019forecasting}:} Near-Collision Set is a large-scale, real-world dataset for near-collision prediction, collected with a stereo camera and LiDAR sensor. In this dataset, color images, accurate depth, and human pose annotations are provided. It contains 13\,658 egocentric video snippets of humans navigating in indoor hallways.

To conclude, Tab. \ref{datasets_list} lists the main datasets we discussed so far, categorizing them according to the input depth map being LR or sparse, as well as reporting the year of collection, the sensor used, the capture environment, the sensed modalities and the number of images. 

\subsection{Objective Functions.}
Notably, the performance of a trained model is directly related to the choice of loss functions, as they estimate the corrections to be applied to weights in the network during training. In this section, we discuss general loss functions for RGB guided ToF imaging, along with loss functions that are frequently employed for the two sub-tasks of GDSR and GDC, respectively.

\subsubsection{General losses}
In RGB guided ToF imaging, the most frequently utilized losses are mean absolute error (MAE), mean square error (MSE), and root mean square error (RMSE). MAE is used in many works~\citep{jaritz2018sparse,shivakumar2019dfusenet,tang2021bridgenet,tang2021joint}, also referred to as $l_{1}$ loss 
\begin{equation}
    \mathrm{MAE}=\frac{1}{N}\sum_{p\in N}\parallel Z_{hq}^{p}-\Hat{Z}_{hq}^{p} \parallel_{1}
\end{equation}
where $\parallel \cdot \parallel_1$ means the $l_1$ norm and $N$ denotes the total number of valid pixels in depth maps. Some methods~\citep{ma2018sparse,zhao2022discrete} adopted MSE ($l_2$ loss) which is defined as
\begin{equation}
    \mathrm{MSE}=\frac{1}{N}\sum_{p\in N}\parallel Z_{hq}^{p}-\Hat{Z}_{hq}^{p} \parallel_{2}
\end{equation}
where $\parallel \cdot \parallel_2$ represents the $l_2$ norm. Another common quantity is RMSE, i.e., the square root of MSE:
\begin{equation}
    \mathrm{RMSE}=\sqrt{\mathrm{MSE}}
\end{equation}

As a perceptual metric for image quality assessment, the Structural Similarity Index (SSIM) is based on the visible structures and is defined as shown in Eq. \ref{ssim}.

\subsubsection{Losses for GDSR.}
To achieve more robust training, a few approaches~\citep{liu2021deformable} use the Charbonnier loss~\citep{charbonnier1994two}:
\begin{equation}
    l_{cha}=\sqrt{\parallel Z_{hq}^{p}-\Hat{Z}_{hq}^{p} \parallel^{2}+\epsilon^{2}}
\end{equation}
where $\epsilon$ denotes a constant that ensures the penalty is non-zero. Sometimes, Peak Signal-to-Noise Ratio (PNSR) is also utilized for GDSR. PSNR can be formulated as follows:
\begin{equation}
    \mathrm{PSNR}=10log_{10}(\frac{Z_{max}^2}{\mathrm{MSE}})
\end{equation}
where $Z_{max}$ is the maximum depth value. The smaller the pixel value difference between the two depth maps, the higher the PSNR.

\subsubsection{Losses for GDC.}
\textbf{Depth loss.} During training, since MAE treats all errors equally and MSE emphasizes the outliers, Huber loss combines the advantages of both:
\begin{footnotesize}
    \begin{equation}   
        l_{huber}=\begin{cases}
        \frac{1}{N}\sum_{p \in N}\frac{1}{2}(Z_{hq}^{p}-\Hat{Z}_{hq}^{p})^2, &|Z_{hq}^{p}-\Hat{Z}_{hq}^{p}|\leq 1 \\
        \frac{1}{N}\sum_{p \in N}(|Z_{hq}^{p}-\Hat{Z}_{hq}^{p}|-\frac{1}{2}), &|Z_{hq}^{p}-\Hat{Z}_{hq}^{p}|>1
        \end{cases}
    \end{equation}
\end{footnotesize}
where $|\cdot|$ is the operator for absolute value. In~\cite{van2019sparse}, another loss named Focal-MSE, inspired by~\cite{lin2017focal}, is proposed and proved to be better than MSE, which is formulated as:
\begin{equation}
    l_{focal}=\frac{1}{N}\sum_{p \in N}(1+\frac{\mathrm{epoch}}{20}|Z_{hq}^{p}-\Hat{Z}_{hq}^{p}|)\cdot(Z_{hq}^{p}-\Hat{Z}_{hq}^{p})^2
\end{equation}

\textbf{Uncertainty-driven loss.} Considering the uneven distribution in the captured depth maps from LiDAR, the uncertainty-driven loss is introduced~\citep{xu2019depth,eldesokey2020uncertainty,zhu2022robust} to focus on more reliable pixels. In their works, GDC is defined as maximizing the posterior probability. The final optimized loss with Jeffrey's prior~\citep{figueiredo2001adaptive} can be written as 
\begin{equation}
    l_{ud}=\frac{1}{N}\sum(e^{-s_n}(Z_{hq}^{p}-\Hat{Z}_{hq}^{p})^2+2s_{p})
\end{equation}
where $s_n$ stands for prediction uncertainty at the $p^{th}$ pixel. 

\textbf{Adversarial loss.} For method adopting a generative adversarial learning strategy~\citep{khan2021sparse,wang2022rgb,nguyen2022patchgan} to perform depth completion, the adversarial loss is needed to discriminate between real and fake images, which can be written as:

\begin{scriptsize}
    \begin{equation}
        l_{adv}\!=\!\min_{G} \max_{D} \mathbb{E}[\log D(Z_{hq})]\!+\!\mathbb{E}[\log (1-D(G(Z_{lq}, I)))]
    \end{equation}
\end{scriptsize}

\noindent where $G(\cdot)$ and $D(\cdot)$ denote the generator and discriminator, respectively.

\textbf{Normal constraint.} 
To further improve depth accuracy, surface normal constraints have been exploited in several works~\citep{zhang2018deep,lee2019depth,xu2019depth}. \cite{zhang2018deep,lee2019depth} compute the depth-normal consistency loss between the predicted normal $R(p)$ at pixel $p$ and the estimated normal $v(p,q)$ from the depth map, as follows:
\begin{equation}
    l_{n1}=\sum_{p,q\in N}\parallel <v(p,q),R(p)>\parallel^2
\end{equation}
where $q$ is a neighbour of pixel $p$, and $<\!\!\cdot\!\!>$ denotes the inner product. To enable models to be trained in an end-to-end fashion, \cite{xu2019depth} propose a negative cosine loss inspired by~\cite{eigen2015predicting}, which is defined as:
\begin{equation}
    l_{n2}=-\frac{1}{N}\sum_{p\in N}R(p)\cdot \Hat{R}(q)    
\end{equation}
where $\Hat{R}(\cdot)$ is the computed normal vector.

\textbf{Smoothness term.} 
Based on the assumption of local smoothness, the smoothness term is introduced in many works~\citep{ma2019self,shivakumar2019dfusenet} to avoid local optimal solutions, penalizing gradients along $x-$ and $y-$ directions. In order to maintain the discontinuities in the final depth maps, some methods~\citep{wong2020unsupervised,wong2021unsupervised,choi2021selfdeco,song2021self,ryu2021scanline,wong2021learning} weight the gradients maps according to the local properties in RGB images:
\begin{equation}
    l_{sm}=\frac{1}{N}\sum_{p \in N}\lambda_{x}(p)|\partial_{x}\Hat{Z}_{hq}^{p}|+\lambda_{y}(p)|\partial_{y}\Hat{Z}_{hq}^{p}|
\end{equation}
where $\lambda_{x}(p)=e^{-|\partial_{x}I^p|}$ and $\lambda_{y}(p)=e^{-|\partial_{y}I^p|}$. 

\textbf{Photometric consistency.} 
In unsupervised methods for GDC, photometric consistency based on the assumption of local smoothness is usually used to refine depth estimation. Given a reference image $I_t$ and its neighboring frame $I_{\tau}$ along the temporal dimension where $\tau\in T=\{t-1, t+1\}$, the goal is to estimate the difference between $I_t$ and its reconstructed version from $I_{\tau}$ as follows:
\begin{equation}
    \Hat{I}_{\tau}(x, \Hat{Z}_{hq})=I_{\tau}(\pi g_{\tau t}K^{-1}\overline{x}\Hat{Z}_{hq}(x))
\end{equation}
where $\overline{x}=[x^T 1]^T$ are the homogeneous coordinates $x\subset \Omega$, $\pi$ denotes the perspective projection, $g_{\tau t}$ is the relative pose of the camera from time $t$ to $\tau$, $K$ represents the camera intrinsic matrix, and $\Hat{Z}_{hq}(x)$ denotes the predicted depth of $x$. 
The relative pose can be either known or estimated by a network.

The photometric consistency is constructed by the combination of $L1$ loss and SSIM on the photometric reprojection as:
\begin{align}
    l_{ph}=&\frac{1}{|\Omega|}\sum_{\tau \in T}\sum_{x\in \Omega}\omega_{co}|I_{t}(x)-\Hat{I}_{\tau}(x)|+ \notag  \\
    & \omega_{st}(1-\mathrm{SSIM}(I_{t}(x),\Hat{I}_{\tau}(x)))
\end{align}
where $\omega_{co}$ and $\omega_{st}$ are weights to balance the two terms. 

\textbf{Pose consistency.}
Given a pair of images ${I_{t}, I_{\tau}}$ as input, a pose network predicts the relative pose $g_{\tau t}\in SE(3)$. When the input of the image sequence is reversed, we expect to obtain $g_t\tau$. The pose consistency loss is formulated as:
\begin{equation}
    l_{pc}=\parallel \log (g_{\tau t}\cdot g_{t\tau})\parallel^{2}_{2}
\end{equation}
where $\log: SE(3) \to se(3)$ is the logarithmic map.

\textbf{Others.}
Finally, we review other commonly used loss functions in Tab.~\ref{fig:loss1}, including depth loss, structural loss, and others.

\begin{table*}[t]	    \scriptsize
	\centering
        \scalebox{0.9}{
	\renewcommand\tabcolsep{4pt} 
	\begin{tabular}{lll}
		\toprule
		\textbf{Types} & \textbf{Notation} & \textbf{Purpose}     \\
		\midrule
              \multirow{3}{*}{\textbf{Depth loss}}  &    $l_{berhu}$   &   Berhu loss is a reversion of Huber loss $l_{huber}$.      \\  
              &  $l_{ce}$  &  Cross-entropy is to measure the validaty of pixels.   \\  
              &   $l_{cyc}$  &   Two $l_{1}$ losses for cycle consistency.  \\  \\
              \textbf{Structural loss} &   $l_{grad}$    &  Gradient between the prediction and GT without weights.     \\ \\	  
		   \multirow{3}{*}{\textbf{Others}}   &  $l_{cpn}$  &  RMSE between the prediction and its reconstruction from CPN.       \\
                &   $l_{cos}$    &   Cosine similarity measures the similarity between the prediction and GT.             \\	
                &  $l_{tp}$   &   MAE between the initial depth and the prediction.  \\
                \bottomrule
	\end{tabular}}
        \caption{Main loss terms deployed for GDSR and GDC.}
	\label{fig:loss1}
\end{table*}

\section{Evaluation}   \label{exp}
In this section, we compare the performance of the state-of-the-art GDSR and GDC approaches on widely used benchmarks. We first present the quantitative results for the two categories of RGB guided ToF imaging. Then, the pros and cons of the methods are analyzed. Notably, we focus more on the most recent methods for RGB guided ToF imaging.

\subsection{Comparison of GDSR methods}
We select representative GDSR methods for the quantitative comparison on Middlebury, NYUv2, RGBDD, and DIML datasets with scaling factors of $4\times, 8\times$, and $ 16\times$. For all the datasets, we follow the recent literature~\citep{tang2021joint,zhao2022discrete,yuan2023recurrent,zhong2023deep} to assess the depth quality according to the RMSE metric. 

Tab.~\ref{tab:comparison_gdsr} shows the comparison of the state-of-the-art GDSR methods. In this experiment, bicubic interpolation is set as the baseline method.
From this table, we can see that although the early deep learning methods~\citep{hui2016depth,li2016deep} are designed with simple structures, they show more powerful performance than traditional methods. Using an encoder-decoder architecture, DJFR~\citep{li2019joint}, DepthSR~\citep{guo2018hierarchical} and PAC~\citep{su2019pixel} design two sub-networks that extract RGB and depth features respectively, and attain better performance compared to models built by several conventional CNN layers. 
In order to improve the representation capability of neural networks, PMBANet~\citep{ye2020pmbanet} design a complex structure and further boost the GDSR performance.
Using a lightweight structure, FDSR~\citep{he2021towards} can still produce accurate results. 
As an alternative solution, the kernel prediction network (KPN) employed by DKN~\citep{kim2021deformable} and FDKN also achieves promising results. Further, \cite{zhong2023deep} employ the KPN to handle inconsistent structures between RGB and depth, which improves generalization to real scenes. 
CUNet~\citep{deng2020deep} develops an optimization-based method and yields results comparable to KPNs. Inspired by these optimization-based methods, DCTNet~\citep{zhao2022discrete} and LGR~\citep{de2022learning} attain better performance. LGR~\citep{de2022learning}, in particular, achieves the state-of-the-art on Middlebury and DIML datasets. 
Unlike most methods that can only perform upsampling at fixed integer scales, GeoDSR~\citep{wang2022learning} learns a continuous representation and achieves strong performance in arbitrary scale depth super-resolution on the Middlebury dataset. 
\revise{With an effective multimodal feature fusion strategy, DSR-EI~\citep{qiao2023depth} attains the best results. However, the disadvantage is that it brings a large computational burden.}

From another perspective, most state-of-the-art methods adopt multi-scale or coarse-to-fine strategies, such as SFG~\citep{yuan2023structure}, JGF~\citep{wang2023joint}, and RSAG~\citep{yuan2023recurrent}. On the one hand, these strategies have achieved impressive results in other tasks, showing their effectiveness in learning feature representations. On the other hand, some methods, such as DCTNet~\citep{zhao2022discrete}, LGR~\citep{de2022learning}, DADA~\citep{metzger2022guided} and SSDNet~\citep{zhao2023spherical}, combine classic optimization algorithms with modern CNNs and demonstrate strong performance. Nonetheless, these models cannot be directly applied to many real scenarios, especially on mobile devices. Hence, efficient and effective approaches need to be further investigated.

\begin{table*}[t] \scriptsize
	\renewcommand\tabcolsep{8pt} 
	\centering
        \scalebox{0.8}{
        \begin{tabular}{@{}lcccccccccccc@{}}
		\toprule
		  \textbf{Dataset} & \multicolumn{3}{c}{\textbf{Middlebury}} & \multicolumn{3}{c}{\textbf{NYUv2}} & \multicolumn{3}{c}{\textbf{RGBDD}} & \multicolumn{3}{c}{\textbf{DIML}} \\
            \cmidrule(r){2-4} \cmidrule(r){5-7} \cmidrule(r){8-10} \cmidrule(r){11-13}
            \textbf{Methods} & $4\times$ & $8\times$ & $16\times$ & $4\times$ & $8\times$ & $16\times$ & $4\times$ & $8\times$ & $16\times$ & $4\times$ & $8\times$ & $16\times$    \\
            \midrule
            Bicubic                          & 2.28 & 3.96 & 6.37 & 4.28 & 7.14 & 11.58 & 2.75 & 4.47 & 6.98 & 1.92 & 3.20 & 5.14    \\ 
		GF~\citep{he2010guided}          & 2.49 & 3.98 & 6.08 & 5.05 & 6.97 & 11.1 & 2.72 & 4.02 & 6.68 & 2.72 & 3.40 & 5.13  \\
            DMSG~\citep{hui2016depth}         & 2.11 & 3.74 & 6.03 & 3.02 & 5.38 & 9.17 & 1.80 & 3.04 & 5.10 & 1.39 & 2.34 & 4.02  \\	
            DJF~\citep{li2016deep}           & 1.68 & 3.24 & 5.62 & 2.80 & 5.33 & 9.46 & 1.72 & 2.96 & 5.26 & 1.39 & 2.49 & 4.27  \\
            DJFR~\citep{li2019joint}         & 1.32 & 3.19 & 5.57 & 2.38 & 4.94 & 9.18 & 1.50 & 2.72 & 5.05 & 1.27 & 2.34 & 4.13  \\
		DepthSR~\citep{guo2018hierarchical}  & 2.08 & 3.26 & 5.78 & 3.49 & 5.70 & 9.76 & 1.82 & 2.85 & 4.60 & 1.40 & 2.23 & 3.75  \\	            
            PAC~\citep{su2019pixel}          & 1.32 & 2.62 & 4.58 & 1.89 & 3.33 & 6.78 & 1.25 & 1.98 & 3.49 & 1.27 & 2.03 & 3.45  \\
            CUNet~\citep{deng2020deep}       & 1.10 & 2.17 & 4.33 & 1.92 & 3.70 & 6.78 & 1.18 & 1.95 & 3.45 & 1.18 & 1.88 & 3.25  \\
		DKN~\citep{kim2021deformable}    & 1.23 & 2.12 & 4.24 & 1.62 & 3.26 & 6.51 & 1.30 & 1.96 & 3.42 & 1.27 & 1.86 & 3.22     \\
		FDKN~\citep{kim2021deformable}   & 1.08 & 2.17 & 4.50 & 1.86 & 3.58 & 6.96 & 1.18 & 1.91 & 3.41 & 1.13 & 1.84 & 3.29     \\
		PMBANet~\citep{ye2020pmbanet}    & 1.11 & 2.18 & 3.25 & \textbf{1.06} & \textbf{2.28} & 4.98 & 1.21 & 1.90 & 3.33 & 1.10 & 1.72 & 3.11     \\
		  FDSR~\citep{he2021towards}       & 1.13 & 2.08 & 4.39 & 1.61 & 3.18 & 5.86 & 1.16 & 1.82 & 3.06 & 1.10 & 1.71 & 2.87     \\
            JIIF~\citep{tang2021joint}       & 1.09 & 1.82 & 3.31 & 1.37 & 2.76 & 5.27 & 1.15 & 1.77 & \underline{2.79} & 1.17 & 1.79 & 2.86     \\
            SVLRM~\citep{dong2021learning}   & 1.11 & 2.13 & 4.34 & 1.51 & 3.21 & 6.98 & 1.22 & 1.88 & 3.55 & 1.19 & 1.93 & 3.49   \\
            AHMF~\citep{zhong2021high}       & 1.07 & 1.63 & 3.14 & 1.40 & 2.89 & 5.64 & 1.10 & 1.73 & 3.04 & 1.10 & {1.70} & 2.83     \\
		DCTNet~\citep{zhao2022discrete}  & 1.10 & 2.05 & 4.19 & 1.59 & 3.16 & 5.84 & {1.08} & 1.74 & 3.05 & {1.07} & 1.74 & 3.09    \\
		LGR~\citep{de2022learning}         & 1.11 & 2.12 & 4.43 & 1.79 & 3.17 & 6.02 & 1.30 & 1.83 & 3.12 & 1.25 & 1.79 & 3.03     \\		
            RSAG~\citep{yuan2023recurrent}     & -- & -- & -- & 1.23 & 2.51 & 5.27 & 1.14 & 1.75 & 2.96 & -- & -- & --     \\            
            SFG~\citep{yuan2023structure}      & -- & -- & -- & 1.45 & 2.84 & 5.56 & -- & -- & -- & -- & -- & --   \\        
            DAGF~\citep{zhong2023deep}         & 1.15 & 1.80 & 3.70 & {1.36} & 2.87 & 6.06 & 1.17 & 1.75 & 3.10 & 1.15 & 1.76 & 3.16     \\  
            JGF~\citep{wang2023joint}          & -- & -- & -- & 1.57 & {2.48} & \textbf{4.11} & 1.25 & 1.85 & 3.03 & -- & -- & --     \\            
            DADA~\citep{metzger2022guided}     & \underline{1.02} & {1.72} & {3.16} & 1.55 & 2.88 & 5.34 & 1.12 & {1.70} & 2.89 & \underline{1.06} & \underline{1.62} & \underline{2.64}     \\
            GeoDSR~\citep{wang2022learning}    & {1.04} & \underline{1.68} & \underline{3.10} & 1.42 & 2.62 & \underline{4.86} & {1.10} & \underline{1.69} & {2.84} & {1.07} & \underline{1.62} & {2.70}     \\
            SSDNet~\citep{zhao2023spherical} & \underline{1.02} & 1.91 & 4.02 & 1.60 & 3.14 & 5.86 & \underline{1.04} & 1.72 & 2.92 & -- & -- & --   \\
             DSR-EI~\citep{qiao2023depth} & \textbf{0.97} & \textbf{1.53} & \textbf{2.32} & \underline{1.21} & \underline{2.46} & 4.95 & \textbf{0.91} & \textbf{1.37} & \textbf{2.10} & \textbf{0.69} & \textbf{1.19} & \textbf{1.96}   \\
            \bottomrule
	\end{tabular}}
        \caption{\textbf{GDSR -- Results on Middlebury, NYUv2, RGBDD and DIML datasets.} The lower the RMSE, the better. Best results in bold, second-best are underlined.}
	\label{tab:comparison_gdsr}
\end{table*}


\subsection{Comparison of GDC methods}
This section gives the quantitative results achieved by several representative methods on the two benchmark datasets, i.e., KITTI driving scenes and NYUv2 indoor scenes datasets, as shown in Table~\ref{tab:comparison_gdc}. For the KITTI dataset, we use RMSE, MAE, iRMSE, and iMAE metrics to evaluate model performance. The results on NYUv2 are measured using RMSE and REL metrics.

From the table, we can see that unsupervised models have been making continuous progress. Even so, their performance still lags significantly behind their supervised counterparts. 
By using uncertainty as auxiliary information for depth, MS-Net ~\citep{eldesokey2019confidence}, pNCNN ~\citep{eldesokey2020uncertainty}, MJPM ~\citep{zhu2022robust} and PADNet ~\citep{peng2022pixelwise} can filter out noisy input and invalid pixels, thus achieving better performance than versions without using uncertainty.
As another auxiliary-based method, multi-task learning has been adopted by various methods to improve generalization performance. The results produced by Multitask GANs~\citep{zhang2021multitask} and SIUNet ~\citep{ramesh2023siunet} demonstrate the effectiveness of this kind of approach.
Networks using dual-branch structure~\citep{tang2020learning,liu2021learning,zhao2021adaptive,gu2021denselidar,chen2023agg,yu2023aggregating} also attain good performance. These methods fully leverage multi-scale features to recover the details of depth maps progressively. Based on the framework, RigNet~\citep{yan2022rignet} obtains the best results on NYUv2.

We also analyze the performance of the affinity-based methods. The affinity matrix produced by these methods, such as NLSPN~\citep{park2020non} and GraphCSPN~\citep{liu2022graphcspn}, are shown to improve depth accuracy through spatial propagation consistently. In addition to spatial propagation, DySPN~\citep{lin2022dynamic} employs the attention mechanism to achieve excellent results. Furthermore, CompletionFormer~\citep{Zhang2023CompletionFormer} combines CNN with a recently popular and dominant technique, Vision Transformer, to achieve better performance in depth completion. 
\revise{
In the most recent work, LRRU~\citep{LRRU_ICCV_2023} outperform other state-of-the-art approaches by learning spatially-variant kernels and updating iteratively.
}

\begin{table*}[htbp] \scriptsize
	\renewcommand\tabcolsep{8pt}
	\centering
        \scalebox{0.78}{
        \begin{tabular}{@{}lcccccccc@{}}
        \toprule
        \multirow{2}{*}{\textbf{Method}} &  \multirow{2}{*}{\textbf{Loss Function}} &  \multirow{2}{*}{\textbf{Learning}} &  \multicolumn{4}{c}{\textbf{KITTI Dataset}} &  \multicolumn{2}{c}{\textbf{NYUv2 Dataset}}  \\
        \cmidrule(r){4-7} \cmidrule(r){8-9}
        &       &       &   RMSE  &  MAE   &   iRMSE  &  iMAE &  RMSE &  REL\\   
        \midrule
        DDP~\citep{yang2019dense}            & $l_{1}+l_{cpn}+l_{SSIM}$     &    Un  & 1263.19     &  343.46     &  3.58     &  1.32 &    --   &   --    \\
        VIO~\citep{wong2020unsupervised}     & $l_{ph}+l_{1}+l_{pc}+l_{sm}$      &    Un   & 1169.97      &  299.41     &  3.56     &  1.20  &   --   &   --    \\
        lsf~\citep{qu2020depth}              & $l_1+l_{SSIM}$      &    Un   &  885.00     & 225.20      &  3.40     & --  & 0.134  & --  \\
        KBNet~\citep{wong2021unsupervised}   & $l_{ph}+l_{1}+l_{sm}$      &    Un  &  1069.47     &  256.76     &  2.95     & 1.02  &   -- &    --      \\
     
        CSPN~\citep{cheng2018depth}               &  $l_{2}$      &    Su   &  1019.64   &  279.46   &  2.93   &   1.15   & 0.117    & 0.016   \\
        MS-Net ~\citep{eldesokey2019confidence}   &  $l_{1}$      &    Su   &  859.22   &  207.77   &  2.52   &   0.92   & 0.117    & 0.016   \\
        Deeplidar ~\citep{qiu2019deeplidar}       &  $l_{2}+l_{n2}$      &    Su   &  758.38   &  226.50   &  2.56   &   1.15   & --    & --  \\
        pNCNN ~\citep{eldesokey2020uncertainty}     &  $l_{2}$      &    Su   &  988.57   & 228.53   &  2.71   &   1.00   & -    & -   \\
        NLSPN~\citep{park2020non}   & $l_{1}$ or $l_{2}$      &    Su   &  741.68    &  199.59     &  1.99     & 0.84   &   {0.092}   &   \underline{0.012}      \\
        GuideNet~\citep{tang2020learning}    &  $l_{2}$      &    Su   &  736.24   &  218.83   &  2.25    &    0.99    & 0.101 & 0.015   \\
        KernelNet ~\citep{liu2021learning}      & $l_{1}+l_{ce}+l_{grad}$     &    Su   & 785.06    &  218.60   &  2.11     &  0.92     &   0.111   &   0.015   \\
        ACMNet~\citep{zhao2021adaptive}         & $l_{2}+l_{sm}$     &    Su   & 744.91    &  206.09   &  2.08     &  0.90  &  0.105    &0.015   \\
        DenseLiDAR~\citep{gu2021denselidar}     & $l_{2}+l_{grad}+l_{SSIM}$     &    Su   & 755.41    &  214.13   &  2.25     &  0.96   &   --  &   --    \\
        RigNet~\citep{yan2022rignet}        & $l_{2}$     &    Su   & 712.66    &  203.35   &  2.08    &  0.90  &   \underline{0.090}   &   {0.013}    \\

        Ssgp~\citep{schuster2021ssgp}   & $l_{2}$      &    Su   &  833.00    &  204.00     &  --     &   --  &   --  &   --    \\
        DenseLConv~\citep{lee2022multi}    &  $l_{2}$      &    Su   &  729.88    &  210.06   &  2.10     & 0.93    &   0.099   &   0.015    \\
        GraphCSPN~\citep{liu2022graphcspn}      &  $l_{1}$      &    Su   &  738.41    &  199.31   &  {1.96}     & {0.84}   &  \underline{0.090} &   \underline{0.012}   \\
        SpAgNet~\citep{conti2023sparsity}       &  $l_{1}$      &    Su   &  844.79    &  218.39   &  -     & -   &  0.114 &   0.015   \\

        Multitask GANs~\citep{zhang2021multitask}   &  $l_{adv}+l_{2}+l_{cyc}$      &    Su   &  746.96   &  267.71   &  2.24    &    1.10    & - & -   \\
        SIUNet ~\citep{ramesh2023siunet}    &  $l_{1}$      &    Su   &  1026.61   &  227.28   &  2.73   &    0.96    & 0.138    & 0.015   \\
       
        FCFR-Net ~\citep{liu2021fcfr}   &  $l_{1}+l_{2}$      &    Su   &  735.81   &  217.15   &  2.20   &    0.98    & 0.106    & 0.015   \\
        MFF-Net ~\citep{liu2023mff}     &  $l_{1}+l_{2}$      &    Su   &  719.85   &  208.11   &  2.21   &    0.94   & 0.100    & 0.015   \\

        MJPM ~\citep{zhu2022robust}     &  $l_{1}+l_{2}$      &    Su   &  795.43   & \underline{190.88}   &  1.98   &   0.83  & -    & -   \\
        PADNet ~\citep{peng2022pixelwise}      &  $l_{1}+l_{focal}$      &    Su   &  746.19   & 197.99   &  {1.96}   &   0.85  & 0.094    & \underline{0.012}   \\
        DySPN~\citep{lin2022dynamic}    & $l_{1}+l_{2}$      &    Su   &  {709.12}    &  {192.71}    &  \underline{1.88}     & \underline{0.82}     &  \underline{0.090}  &   \underline{0.012}     \\
        CompletionFormer~\citep{Zhang2023CompletionFormer}      &  $l_{1}+l_{2}$      &    Su   &  \underline{708.87}   &  203.45   &  2.01     & {0.84} & \underline{0.090} & \underline{0.012}   \\
        AGGNet~\citep{chen2023agg} & $l_2+l_{huber}$ & Su & -- & -- & -- & -- & 0.092 & 0.014 \\
        PointDC~\citep{yu2023aggregating}  & $l_1+l+2+l_{grad}$ & Su & 736.07 & 201.87 & 1.97 & 0.87 & \textbf{0.089} & \underline{0.012}    \\
        LRRU~\citep{LRRU_ICCV_2023}  &  & Su & \textbf{696.51} & \textbf{189.96} & \textbf{1.87} & \textbf{0.81} & 0.091 & \textbf{0.011} \\
        \hline
        \end{tabular}}
        \caption{{GDC -- Results on KITTI and NYU-v2 datasets. \textbf{Un} and \textbf{Su} indicate unsupervised and supervised models.} The lower the RMSE, the better. Best results in bold, second-best are underlined.}
	\label{tab:comparison_gdc}
\end{table*}

\section{Discussion and future trends}  \label{future}
Although some reviewed methods have achieved promising results on RGB guided ToF imaging, some issues hinder their practical application. Here, we mainly describe the challenges and future trends from the following four aspects.

\subsection{Unsupervised methods with lightweight structure}
As described before, most methods perform GDSR and GDC after training with a supervised paradigm. However, labeled data is usually difficult to collect. In addition, the performance of mass-produced sensors generally fluctuates to a certain extent, and it is unrealistic to fine-tune the model on data collected from any existing sensor. Therefore, unsupervised learning methods need to be developed without ground truth depth. In particular, state-of-the-art unsupervised learning methods cannot achieve the performance of supervised learning methods, so there is still much room for improvement in this direction.

On the other hand, these methods often require many parameters and much computation to achieve better results, which makes the models unable to run on many devices at high speed, especially widely used mobile devices. While transferring the information to the cloud for processing is one solution to this problem, it raises thorny privacy issues. Thus, efficiently recovering high-quality depth is necessary and remains an open challenge. One recent example is SeaFormer~\citep{wan2023seaformer}, which proposes a versatile mobile-friendly backbone based on the vision transformer. Developing a simple yet efficient unsupervised model for RGB guided ToF imaging is challenging but highly desirable.

\revise{
\subsection{Cross-domain generalization}
Another issue concerns the ability of networks to generalize to new domains. This topic has been mostly overlooked in the literature despite its relevance for the widespread diffusion of depth estimation in practical applications. Purposely, \cite{Bartolomei_2024_3DV} proposed a method to improve cross-domain generalization for depth completion by exploiting the virtual projection pattern paradigm proposed in \cite{Bartolomei_2023_ICCV} and casting depth completion as a correspondence problem using a virtual stereo setup. Processing properly hallucinated fictitious stereo pairs with a robust stereo network that possibly exploits the RGB image context, such as RAFT \cite{RAFT_STEREO}, leads to more robust cross-domain generalization than state-of-the-art depth completion networks.  
}

\subsection{ToF with varifocal lens color camera}      
RGB cameras in many devices, such as mobile phones and drones, have variable focal lengths, while ToF cameras usually employ fixed focal length lenses. Different focal lengths correspond to different depths of field and angles of view. When a ToF camera captures a distant object, boundaries are usually blurred. If the color image autofocuses on the object, it can provide rich structural details and transfer to the depth map. Nonetheless, it also brings a problem that the scales of the images from the two modalities may differ. In this case, the RGB image and the depth map cannot be directly aligned by the registration technique, i.e., a new calibration is required. Hence, further investigation needs to be conducted on how to align two scale-inconsistent images, i.e., the RGB image and the depth map, and simultaneously transfer the scene geometry in the RGB image to the depth map.

\subsection{Spot-ToF}    
The main disadvantage of ToF cameras using floodlight projection, currently the most commonly used technique in mobile devices, is their close operating distance. The scheme proposed by \cite{achar2017epipolar} can significantly extend the measurable distance of the ToF camera, but the long acquisition time of each image limits the practical application. 
On the other hand, direct ToF (d-ToF) can capture distant objects, whereas the resolution of the sensors is much lower than that of indirect ToF (i-ToF) with tens of thousands of pixels, e.g., $8\times 8$ or $24\times 24$. 
LiDAR also faces the same problem, and its manufacturing cost is too expensive, making it currently unable to be used as a consumer-grade camera.
\cite{ge2021tof} provide an alternative to alleviate the issue. Specifically, they replaced the optical diffuser in the ToF camera projector with a diffraction optical element (DOE), which is used to evenly distribute the incident light into thousands of laser beams. In this way, each laser beam can have a measurable distance greater than that of floodlight illumination. However, this approach faces two challenges. 
First, a laser beam forms a spot in the sensor, corresponding to several pixels. The uneven intensity distribution of the spot may lead to inconsistencies in the measurement, so it is necessary to develop new depth correction methods to obtain an accurate depth map. 
Second, the captured images are sparse (similar to LiDAR but at a lower cost), so effective and efficient algorithms need to be developed to estimate dense depth under this setting.

\subsection{Under-display RGBD camera}
In recent years, in pursuit of a better visual experience, full-screen devices have attracted the attention of industry and academia. 
This fact also brings a major challenge: images captured by under-display cameras are severely degraded, requiring effective strategies for restoration.
Most of the existing work is to study the image restoration of under-display RGB cameras (UDC), while few works focus on under-display ToF (UD-ToF) depth restoration. 

\textbf{Under-display camera.}
Recently, smartphones with full-screen displays, eliminating the need for bezels, have become a new product trend and motivated manufacturers to design a brand-new imaging system, the Under-Display Camera. Placing a display in front of the camera lens can increase the screen-to-body ratio for a better user experience. However, the broad commercial manufacturing of UDC is prevented by poor imaging quality due to the inevitable degradations, such as blurs, noise, diffraction artifacts, color shifts, etc.

UDC image restoration has been studied in a few literary works. As the pioneered of this work,~\cite{zhou2021image} devise a Monitor Camera Imaging System (MCIS) to collect paired data and offer two practical solutions for UDC image restoration, including a deconvolution-based Wierner Filter pipeline and the data-driven method.~\cite{yang2021designing} redesign the display layout to improve the quality of recovered images.
To address spatially variant blur and noise,~\cite{kwon2021controllable} develop a controllable image restoration algorithm, which performs well on both a monitor-based aligned dataset~\citep{zhou2020udc} and a real-world dataset.~\cite{feng2021removing} propose a domain-knowledge-based network to restore the UDC images. However, this work requires a point-spread function (PSF) as a prior, thus failing to work well when the PSF is unavailable. To remedy this,~\cite{koh2022bnudc} design an effective two-branched network for UDC image restoration from the perspective of low-frequency and high-frequency information learning. The UDC image restoration methods need paired image datasets that perform poor generalization capability for real degradations. The reason behind this phenomenon is the domain discrepancy between synthesized data and real-captured data. To alleviate this problem,~\cite{feng2023generating} employ collected non-aligned data for UDC image restoration and improve the robustness of the restoration network in real-world scenarios.

\textbf{Under-display ToF.}    
Currently, only one work~\citep{qiao2022depth} has investigated image restoration with a ToF camera placed under a display. In this paper, they propose a cascaded network to restore the depth in a coarse-to-fine manner. Differently from previous methods, in the first stage, they design a complex-valued neural network to recover the ToF raw measurements. In the second stage, they refine the depth based on the proposed multi-scale depth enhancement block. Moreover, to enable the training, they introduce real and synthetic datasets for UD-ToF imaging, respectively. Specifically, the large-scale synthetic dataset is created by analyzing noise in magnitude and phase.

If a color camera is located under a display with a ToF camera, there is greater potential for obtaining high-quality depth maps.  However, the design of models to jointly compensate for multi-modal image degradation is still challenging.

{
\subsection{\revise{Hybrid models}}
Typically, RGBD-based imaging can be addressed by either conventional or deep-learning methods, each of which possesses advantages and disadvantages. A major advantage of conventional methods is that, through well-interpretable hand-crafted models, the predicted depth can be guaranteed to be loyal to the source. However, designing a model with excellent feature representation capabilities is usually extremely challenging. In contrast, deep-learning methods can perform very well on a specific task with sufficient and representative data for training. On the other hand, when there is insufficient data or a biased distribution, these trained models may not achieve the desired results during testing. Therefore, combining the advantages of both to design hybrid models is one of the future trends in RGB guided ToF camera imaging. In particular, the hand-crafted prior, complemented by the powerful feature representation capability of neural networks, can make it easier to achieve a lightweight approach.

\subsection{Online alignment of RGB and depth}
In previous work, RGB and depth maps were usually assumed to have been well-aligned in spatial by calibration operations. However, in practice, the intrinsic ${c_x,c_y}$ and extrinsic parameters ${t_x,t_y}$ change when the camera is deployed on a device or subjected to vibration, which requires online weakly calibration for the RGB guided ToF camera. 
On the other hand, cross-modal online alignment in the time dimension is also necessary for practical applications, as different sensors may not achieve completely accurate synchronization in sensing the environment. Especially when the acquisition is out of sync during motion, the imaging quality of the RGBD camera will be seriously degraded. 
Although some previous efforts~\citep{qiu2019deep} have been devoted to cross-modal correspondence matching, the algorithms consume a long inference time, so temporally and spatially efficient alignment methods remain to be further explored.

\subsection{Lightweight ToF}
To promote the application of ToF cameras in various consumer electronic devices, lightweight ToF cameras~\cite{li2022deltar,liu2023multi} were launched on the market and have received the favor of many manufacturers for their low power consumption, low cost, and compact structure. 
Unlike common ToFs, lightweight ToF cameras generate measurements as depth distributions rather than specific values.
Specifically, this type of camera can be used in many fields, such as AR/VR and obstacle avoidance.
Nonetheless, the lightweight design inevitably brings drawbacks such as extremely low resolution (e.g., $8\times 8$), which leads to a degradation in the depth quality and cannot support applications requiring high-quality depth. Therefore, using other modal information, such as RGB or IR, to improve the resolution of lightweight ToF depth becomes one of the future trends.

\subsection{Multi-frame processing}
Currently, most algorithms deal with RGB guided ToF imaging in a single-frame fashion, using one degraded depth map as input to infer the corresponding high-quality output. However, the format we usually capture with cameras is a data stream (video) rather than a single image. Therefore, it is crucial to consider the temporal correlation of data and output stable depth maps. There have been some works~\cite{wronski2019handheld,patil2022multi} employing multi-frame processing technology for image denoising, enhancement, and super-resolution, but only a few works~\cite{sun2023consistent} have applied it to ToF depth improvement. Based on the above analysis, fusing ToF depth and RGB images with multi-frame processing technology is a promising direction.

}
\section{Conclusion}  \label{conclu}
In this paper, we have presented a survey of RGB guided ToF imaging methods based on deep learning. Our review covers preliminaries, evaluation metrics, network design, learning protocols, benchmark datasets, and objective functions. According to the measurable distance and the purpose of the ToF camera, we roughly divide the problem faced into two categories: guided depth super-resolution and guided depth completion. Moreover, we collect a quantitative comparison of the surveyed methods on widely used benchmarks and analyze their performance and respective characteristics. Finally, we summarize the current challenges in practical applications as well as promising future trends.

\section*{Acknowledgments}
This work is supported by the National Natural Science Foundation of China (No. 62088102, No. 62376208), Fundamental Research Funds for the Central Universities (No. xzy022023107), China Telecom Group Corporation-Xi'an Jiaotong University Jointly Established Intelligent Cloud Network Science and Education Integration Innovation Research Institute (No. 20221279-ZKT03).

\bibliography{sn-bibliography}

\end{document}